\documentclass[sigconf,nonacm]{acmart}

\usepackage{subcaption}
\usepackage{pifont}

\newcommand{\cmark}{\ding{51}}
\newcommand{\xmark}{\ding{55}}

\AtBeginDocument{%
  }

\setcopyright{none}

\usepackage{etoolbox}

\makeatletter
\apptocmd{\@mktitle}{\vspace{1cm}}{}{}

\pretocmd{\@printtopmatter}{%
  \global\setbox\mktitle@bx=\vbox{%
    \unvbox\mktitle@bx
    \vspace{1cm} 
  }%
}{}{}

\patchcmd{\@typeset@author@bx}
  {\@authorfont\@currentauthors\par\@affiliationfont}
  {\@authorfont\@currentauthors\par\vspace{1.5mm}\@affiliationfont} 
  {}{}
\makeatother

\begin{document}

\title{Compliance Rating Scheme: \\ A Data Provenance Framework for Generative AI Datasets \vspace{0.5cm} }





\author{Matyas Bohacek}
\authornote{Both authors contributed equally to this research.}
\email{maty@stanford.edu}
\orcid{0000-0001-8683-3692}

\affiliation{%
  \institution{Stanford University}
  \streetaddress{}
  \city{}
  \state{}
  \country{}
  \postcode{}
  }

\author{Ignacio Vilanova Echavarri}
\authornotemark[1]
\email{i.vilanova21@imperial.ac.uk}
\orcid{0009-0006-1442-3591}

\affiliation{%
  \institution{Imperial College London}
  \streetaddress{}
  \city{}
  \state{}
  \country{}
  \postcode{}
}

\renewcommand{\shortauthors}{Bohacek and Vilanova}

\begin{abstract}
Generative Artificial Intelligence (GAI) has experienced exponential growth in recent years, partly facilitated by the abundance of large-scale open-source datasets. These datasets are often built using unrestricted and opaque data collection practices. While most literature focuses on the development and applications of GAI models, the ethical and legal considerations surrounding the creation of these datasets are often neglected. In addition, as datasets are shared, edited, and further reproduced online, information about their origin, legitimacy, and safety often gets lost. To address this gap, we introduce the Compliance Rating Scheme (CRS), a framework designed to evaluate dataset compliance with critical transparency, accountability, and security principles. We also release an open-source Python library built around data provenance technology to implement this framework, allowing for seamless integration into existing dataset-processing and AI training pipelines. The library is simultaneously reactive and proactive, as in addition to evaluating the CRS of existing datasets, it equally informs responsible scraping and construction of new datasets.
\end{abstract}




\maketitle

\section{Introduction}
\label{sec:introduction}

As Generative Artificial Intelligence (GAI) applications become increasingly intuitive and their results more realistic, their adoption is becoming widespread \cite{oppenlaender_2023_text}. This exponential growth in performance and adoption in recent years is partly facilitated by the abundance of large-scale open-source datasets, which are often created through unrestricted and opaque data collection practices~\cite{birhane_2021_multimodal, hutchinson_2020_TowardsAF}. Datasets play a crucial role in the AI ecosystem \cite{koch_2021_reduced} as they are the primary source of training for most AI systems. While much of the literature focuses on the development and applications of GAI models, many ethical and legal considerations surrounding dataset creation currently remain unaddressed~\cite{birhane_2021_multimodal, hutchinson_2020_TowardsAF}.

Many AI researchers and practitioners obtain training data on the internet~\cite{rajbahadur2021can,elazar2023s}. With thousands of publicly available datasets~\cite{birhane2021multimodal,wiki:datasets}, platforms like Hugging Face~\cite{lhoest2021datasets} and GitHub~\cite{cosentino2016findings} have become the backbone of today's AI infrastructure. This model of dataset sharing emerged organically~\cite{denton2021genealogy} in the 2000s with pioneering datasets such as ImageNet~\cite{deng2009imagenet} and has been present for several years with no oversight or formal framing~\cite{khan2022subjects,shope2021lawyer}. When the recent boom of GAI erupted in 2023, the same model of dataset sharing practices prevailed---and with it, the same legal and ethical challenges~\cite{martinez2023towards}.

The democratization of GAI has equally caused a surge in malicious activity~\cite{birhane_2021_multimodal,schramowski_2023_safe}, notably through impersonations, copyright infringement, and deepfake pornographic footage~\cite{birhane_2021_multimodal}. Yet, the complexity and opacity of the structure of advanced GAI models result in a lack of traceability and accountability of the datasets that fuel them. To illustrate this challenge, we pose the following example: When a researcher or a practitioner wants to use a publicly available dataset for AI training, standard practice suggests the use of a publicly available image dataset. While the dataset's license is valid, and the use case permitted, much of the dataset is often scraped without the consent of its content creators. Consequently, while the researcher accepts the terms of use outlined by the dataset's authors, and be under the impression that the use of this dataset is legitimate, this might not always be the case. Indeed, for large datasets with millions (and more often than not, billions) of data points are required to train advanced GAI models, and thus, manually inspecting each data point becomes virtually impossible. Yet this common practice might result in a series of grave legal and ethical consequences, varying from copyright infringement, to using illegal material such as Child Sexual Abuse Material (CSAM) --- unknown to the researcher.


This scenario was the case of the LAION-5B dataset~\cite{schuhmann_2022_laion}, powering popular AI image generators such as Stable Diffusion~\cite{rombach2022high}. This dataset ultimately had to be removed from distribution as a result of the two issues described above~\cite{thiel2023generative}.

There are two critical moments that have led to this undesirable outcome. The first pertains the researcher's review of the license and acceptance of the dataset's terms of use. In this largely unregulated landscape~\cite{andreotta2022ai,carter2020regulation}, where copyright infringement remains a contentious issue~\cite{chu2024protect,levendowski2018copyright}, licenses and terms of use present some of the few recognized legal standards~\cite{arora2023value,kusak2022quality,khan2022subjects}. However, as the licenses are often written directly by the dataset authors, users can become overwhelmed and misled by the licensing requirements~\cite{heger2022understanding}. Moreover, a recent study revealed that nearly half of popular AI training datasets exhibit similar issues: they include data whose creators were not asked or informed about its inclusion, potentially violating copyright, while the license makes it look like the use of the data is permissible~\cite{longpre2023data}. As ethical and legal frameworks for AI datasets are still in their infancy~\cite{verhulst2023urgent,labastida2020licensing}, it is not clear who is responsible for such misconduct. Currently, the responsibility is often placed on the authors of the datasets~\cite{Li_2023}. Nevertheless, the dataset license agreed upon by the researcher could put the liability on them as well~\cite{zirpoli2023generative}. Depending on the jurisdiction and context, this could render the researcher liable instead of the dataset author.

The second critical moment involves the researcher's inability to verify the dataset authors' claims of ethical and legal data sourcing. The researcher has no alternative but to trust the dataset authors on this (i.e., a trust-based system). From an ethical and legal perspective, these two key moments beg the following questions: first, on the side of the dataset authors, can the practice of including unauthorized data points during scraping be prevented? Second, on the side of the researcher, can the accuracy of the presented license and dataset policies be verified to prevent misuse?

To the best of our knowledge, current literature and practice shows that no. The only way for the dataset author to verify that all scraped data points satisfy their chosen criteria is to manually inspect each one. The same limitation applies to the inverse case of a user testing whether the dataset's self-reported license and policies match reality.

To address these critical questions, we propose a set of four practical principles for accountable, license-compliant datasets. These principles are informed by existing dataset sharing practices and latest data provenance technologies. We then conceptualize the Compliance Rating Scheme (CRS) as a trustless tool to evaluate a given dataset's compliance with these principles. Finally, we develop a Python library, \textit{DatasetSentinel}, which allows dataset authors to integrate these principles into their scraping pipelines and enables users to verify the CRS of datasets they are considering. We open-source the library at \url{[anonymized]}.

The rest of the paper is outlined as follows. We first contextualize GAI within the broader context of datasets for AI training and provide an overview of current concerns. We then position these concerns within the existing ethical and legal frameworks, which we synthesize into the practical principles that guide our solution. This discussion transitions into a description of the CRS. Next, we introduce the development, structure, and evaluation of \textit{DatasetSentinel}, the Python library we open-source. Finally, we discuss the implications, limitations, and future directions of this research.

\section{Background and Related Work}

In this section, we provide an overview of the GAI landscape, focusing on training datasets and their misuse. We identify a gap in existing work, into which we later position our contribution.

\subsection{Generative AI}

Generative AI (GAI)~\cite{banh_2023_GenerativeAI} refers to AI systems that can synthesize novel text~\cite{zhao_2023_ASO}, image~\cite{bie_2023_RenAIssanceAS}, video~\cite{singh_2023_ASO,liu_2021_AIEmpoweredPV}, audio~\cite{zhang_2023_ASO}, and other modalities. These systems have recently undergone a substantial leap in generation quality and ease of use~\cite{maslej_2023_ArtificialII}. We first saw this in the domain of text generation with the advent of large language models (LLMs)~\cite{brown_2020_LanguageMA,touvron_2023_LLaMAOA}. Based on a simple premise of completing the next words in sentences, LLMs have improved to the point where they evince emergent capabilities~\cite{wei_2022_EmergentAO,valmeekam_2023_OnTP}, such as text analysis and question answering. In fact, the quality of AI-generated text today is such that humans, in some cases, cannot distinguish AI-generated texts from human-written texts~\cite{bohacek_2023_TheUA,casal_2023_CanLD}. 

Similar leaps are being made on the front of image, audio, and video generation~\cite{zhang2023text}. It was not long ago when text-to-image generation was restricted to Generative adversarial networks (GANs)~\cite{goodfellow_2014_GenerativeAN}, which could only generate domain-specific images~\cite{karras_2018_ASG}. Today, diffusion-based models~\cite{zhang_2023_TexttoimageDM} work across a wide range of subject domains and create text-conditioned images of high quality. Similar to text, the latest methods for image generation got to a point where, in some contexts, humans cannot distinguish AI-generated images from real photos~\cite{nightingale_2022_AIsynthesizedFA}. Methods for video~\cite{wang_2023_VideoComposerCV,wu_2022_TuneAVideoOT,khachatryan_2023_Text2VideoZeroTD,singer_2022_MakeAVideoTG} and audio~\cite{natsiou_2021_AudioRF,kalchbrenner_2018_EfficientNA,xue_2024_AuffusionLT} generation are a more recent addition to the scope of modalities generated by AI, but we can expect them to follow a similar trajectory to the text and image modalities.

\subsection{Datasets}

The scale and quality of AI training datasets have been essential to the recent leap in GAI systems~\cite{halevy_2009_TheUE,maslej_2023_ArtificialII}. As such, datasets play an essential role in the AI ecosystem~\cite{koch_2021_reduced} because they are the primary source of training for most AI systems. That is because---despite advances in reinforcement learning and other modes of AI---supervised training of AI models, in which a model is trained once before put to use, still dominates. Beyond training, datasets also allow teams to compare the performance of their solutions against a standardized benchmark. More broadly, datasets steer the focus and work within the community~\cite{scheuerman_2021_DoDH}. 

Albeit significant progress has been made in the areas of foundation models, fine-tuning, and knowledge transfer, most AI systems require task-specific datasets to achieve good performance~\cite{villalobos_2022_WillWR, halevy_2009_TheUE}. Therefore, new datasets are constantly being released by research institutions~\cite{paullada_2020_DataAI}, companies~\cite{polyzotis_2018_data}, and laypeople~\cite{chiang_2022_ExploringTE} alike as new AI tasks and contexts emerge. Notably, the data acquisition and annotation of a large dataset is financially demanding~\cite{cong_2021_DataPI} and, while some companies can afford to undergo such a project with manual curators and annotators, many entities resort to a less expensive data acquisition mode through internet scraping~\cite{khder_2021_WebSO}. Many of today's datasets for AI training are thus indiscriminately scraped from the internet~\cite{paullada_2020_data, chugunkov_2018_creation}, often with little or no consideration of the ethical and legal implications of such practice.

\subsection{Dataset Life Cycle and Stakeholders}

We consider the distinction between the dataset author and AI practitioner in the dataset life cycle essential, as both parties approach it with different incentives and risks~\cite{heger_2022_UnderstandingML}. As such, we frame the life cycle of a dataset by its creation (performed by the dataset author) and by its use (performed by the AI practitioner). The author of the dataset (e.g., a research institution, a company, or an individual) first identifies the scope of the dataset: they select which task(s) and context(s) the dataset will address and create fundamental policies about the ingested data~\cite{hutchinson_2020_TowardsAF, sambasivan_2021_EveryoneWT}. Next, they set up the data ingestion sources (e.g., custom capture, data purchase, or internet data scraping) and annotation mechanisms (e.g., hiring annotators or employing automated solutions) and, finally, proceed with the construction itself. Once constructed, the dataset's license and terms of use are packaged with the dataset and shared. Over time, the owner may decide to make changes to the dataset. Such modifications may include simple error fixes or, if the changes are significant enough to justify so, constitute a new version of the dataset. As \textit{Hutchinson et al.} argue, datasets powering AI are often used, shared, and reused with little visibility into the processes of deliberation that lead to their creation~\cite{hutchinson_2020_TowardsAF}. 

The paths of the dataset author and the AI practitioner intersect at the dataset distribution platform. Often, the datasets are shared on Hugging Face Datasets\footnote{\url{https://huggingface.co/datasets/}}, Kaggle\footnote{\url{https://www.kaggle.com}}, GitHub\footnote{\url{https://github.com}}, or custom websites. At this point, the AI practitioner is considering which datasets would best fit their use case~\cite{zhang_2023_navigating}. Once they decide which dataset(s) to use, they obtain the data from the platform and proceed to the model training, evaluation, and potential deployment (inference).

\subsection{Challenge to Address}

Manually inspecting every data point included in a dataset is virtually impossible as it is not uncommon for a single dataset to contain millions to billions of such data points. This makes it challenging for dataset authors to filter incoming data and for AI practitioners to verify whether the contents of a dataset match its description. There needs to be a systematic trustless approach to infer the provenance of a single data point that would enable the dataset author to filter incoming data points effectively and AI practitioners to assess a considered dataset. Existing metadata (e.g., EXIF) sometimes includes relevant information about the license, author, AI opt-out, etc., but this information is often missing, and even when present, it is inconsistent in formatting and terminology. There needs to be a systematic approach to address data provenance, in which license, AI training consent, and other preferences would be automatically embedded.


Moreover, the abundance of large-scale open-source datasets derives from the legal vacuum of online data collection and use practices. This state of affairs has led to calls for a more responsible, transparent, accountable, and human-centered approach to AI dataset practices~\cite{birhane_2021_multimodal,hutchinson_2020_TowardsAF}. Consequently, we argue that a new framework is needed to better these unrestricted and unaccountable practices. We identify the modalities of image, video, and audio datasets as the most pressing to address. These modalities constitute some of the most prominent kinds of data points in datasets for GAI training and, as mentioned above, can, in some contexts, pose imminent privacy concerns for individuals.

\subsection{Dataset Principles}

The literature on dataset ownership encompasses a wide range of sub-themes, such as privacy, security, stewardship and governance, and transparency~\cite{asswad_data_2021}. Yet, there is no standardized definition of what such ownership entails~\cite{asswad_data_2021}. This begs the question of whether users' data should be considered some sort of property, and whether this status would change when aggregated to a dataset protected under Intellectual Property (IP) laws. Legal theories suggest that data (information) cannot be owned~\cite{hummel_own_2020}. Common law does not recognize property in facts or information and considers data as such. Continental (Civil) law follows a similar approach but presents data rights as an “extension or subset of fundamental human rights”~\cite{hummel_own_2020}, which is also unsuitable for propertization and commercialization. For this reason, most legal scholars focus on the protection of data instead~\cite{cofone2021beyond}. However, as we have discussed, individuals have little to no practical rights on how their personal data is used in the context of AI training datasets.

In the case of AI, applications involve primarily two categories of tort: dignitary and property. The distinction derives from the nature of the harm caused. Dignitary torts typically encompass harm to a person's reputation, honor, or dignity, such as unauthorized use of personal images – like pornographic deepfakes. Conversely, property torts in AI contexts often relate to interference with one's property rights - such as copyright infringement with artists' work used to train GAI models. While legal categorization is relatively straightforward, successful prosecution and liability remain incredibly complex.

We break down this complexity into three main areas. First, data collection and dataset practices: AI models often use publicly available data (such as photos, videos, and voice samples) scraped from the internet to generate context-specific outcomes. These data collection practices are often unrestricted and opaque, and rarely have explicit consent for specific uses~\cite{adjerid_choice_2019, draper_corporate_2019, solove_introduction_2012, solove_myth_2021}. Second, overlapping jurisdiction and areas of law): Part of the legal complexity and ambiguity derives from a series of overlapping legal areas, such as artistic freedom, freedom of expression, the right to information, the right to privacy, and personality rights, among others. Furthermore, different jurisdictions might interpret these rights differently. Third, there is a lack of liability and accountability. The anonymity of the internet makes it difficult to determine who created or distributed the AI application’s output (e.g., a deepfake).

While the data coming out of AI systems has been an active area of study (from deepfake detection~\cite{bohacek_protecting_2022_arxiv,bohacek_protecting_2022_pnas} to watermarking to tracing detailed provenance information of AI-generated content~\cite{rosenthol_2020_content,rosenthol_2022_c2pa}), the data coming into these systems during training, which leads to AI and dataset misuse, has not~\cite{bhardwaj2024machine,ajmani2024data}. Modern data protection laws such as the EU’s General Data Protection Regulation (GDPR) (2018)~\cite{noauthor_general_2018} and the California Consumer Privacy Act (2018)~\cite{noauthor_california_2018} are built on The Fair Information Practice Principles (FIPPS) published by the Organization for Economic Cooperation and Development (OECD) in 1980~\cite{noauthor_apec_2004,noauthor_oecd_1980}, and has ever since been the guiding model for data protection. These principles have been adopted through various institutions and improved through frameworks such as the EU Data Protection Directive Principles (1992)~\cite{noauthor_eur-lex_1992}, the Federal Trade Commission Privacy Principles (1998)~\cite{noauthor_privacy_1998}, and the Asia-Pacific Economic Cooperation Privacy Framework (2004)~\cite{noauthor_apec_2004}. Yet, as discussed, the Fair Information Practice Principles and most of the laws derived from them have failed in practice~\cite{cate_failure_2006}, as the data protection regimes built on them come short in providing a high standard of effective and efficient data protection and use~\cite{cate_failure_2006}. As such, data protection is not an end in itself, but rather a tool for enhancing individual and societal welfare. We aim to pursue this goal by proposing an initial set of four practical principles to consider for dataset compliance in the context of AI. Inspired by prior work and data protection laws, these four principles are designed to be technologically implementable, and to provide actionable measures for prosecution in the eventuality of misuse. These principles consist of:

\begin{enumerate}
    \item Responsibility and Liability 
    \item Effective and Efficient Enforcement
    \item Prevention of Harm 
    \item Transparency and Fair Use
\end{enumerate}

\subsection{Data Provenance}

Data provenance~\cite{pan_2023_DataPI} refers to the records about the origin, ownership, and evolution of a file. It is concerned with any relevant information from the moment the file was created---be it as an authentic recording or as a synthetic digital product---to its present form. This information includes details about the entities, software, and specific changes, if applicable, that have in any way manipulated the file from its inception~\cite{magagna_2020_data,werder_2022_EstablishingDP}. Moreover, the data provenance may capture additional information about its author's decisions for sharing it with third parties, including the license under which it is shared, whether or not it may be included in AI training, etc.

Establishing data provenance for files disseminated over the internet may be challenging~\cite{chen_2017_DataPA}, especially as they may be stripped of their basic metadata or additional attachments. Therefore, the literature has studied cryptographic methods for provenance~\cite{engram_2021_ProactivePP}, which allow for verifying any assertions made about a file in its provenance metadata. While there are many contexts in which establishing data provenance may be essential, this technology has gained most of its recognition recently amidst a wave of fake AI-synthesized images on the internet~\cite{sidnam_2022_UsableCP}.

The Coalition for Content Provenance and Authenticity (C2PA) data provenance specification~\cite{rosenthol_2022_c2pa} created a standardized framework for data provenance metadata. As of yet, this is the largest effort striving to establish a standardized approach to deployable data provenance on the internet, and it has received traction from industry and academia alike. Content Authenticity Initiative (CAI) then materialized this standard into a functional, cryptography-based library and metadata scheme~\cite{rosenthol_2020_content}.

\section{Compliance Rating Scheme}
\label{sec:crs}

Our contribution comprises two parts, the first is the Compliance Rating Scheme (CRS). It is a set of criteria and a summarizing score that together serve as an intuitive indicator of a given dataset's compliance with the principles outlined above. The CRS score is evaluated based on the following six criteria:

\begin{enumerate}
    \item The sourcing, filtering, and pre-processing employed during data acquisition and annotation of the dataset is transparent. The code for these processes is either fully open-sourced or is described at a level of detail that would enable full reproduction of the dataset.
    \item The dataset complies with the license and allowed use described in the provenance metadata of each included data point. This means that the licenses and allowed use of each individual data point fall within the scope and allowed use of the dataset as a whole.
    \item The dataset flags any data points where compliance with the provenance metadata is inconclusive.
    \item The dataset has an opting-out mechanism, allowing authors of the included data points to request their removal from the dataset if they had not previously given consent.
    \item Any changes made to the content of the dataset---both to the data points themselves and their annotations---are traceable. There is a designated trace log that includes dated records of changes, listing which data points were impacted and how.
    \item The dataset adds the dataset source and the retention period into the provenance metadata of each included data point.
\end{enumerate}

The CRS score summarizes the dataset's compliance with these criteria into a letter on the scale from "A" (the highest, most compliant score) to "G" (the lowest, least compliant score). Starting at "G", each satisfied criterion moves the CRS of the evaluated dataset up by one letter grade. This means that, if a dataset does not meet any of these criteria, it receives a CRS of "G". Contrarily, if the dataset meets all criteria, it receives a CRS of "A".

While there are many contexts in which this assessment could be desired, it is primarily targeted at AI practitioners when they are deciding which dataset(s) to use for training in their AI project. Returning to our example of an AI researcher from Section~\ref{sec:introduction}, the researcher can benefit from the CRS score to determine which datasets out of the ones she was considering satisfy the legal and ethical standards she desired. Even if a dataset's description claimed so, she could verify that through the CRS score, and thus remove the element of trust in the dataset's creator good faith from the equation.
 



 
\section{Library}

The second part of our contribution is \textit{DatasetSentinel}, an open-source Python library implementing the CRS. The library, available at \url{[anonymized]}, is written in Python, the most popular programming language for AI research and development~\cite{gonzalez_2020_state}. It can be easily integrated into existing dataset and AI pipelines as it is compatible with PyTorch~\cite{Paszke_2019_PyTorch}, TensorFlow~\cite{Abadi_2016_TensorFlow}, MLX~\cite{hannun_2023_mlx}, HuggingFace~\cite{wolf_2019_HuggingFace}, Kaggle, and custom dataset-sharing platforms, requiring minimal changes to existing code structures.

The library leverages the Content Authenticity Initiative's (CAI) library~\cite{rosenthol_2020_content} and the Coalition for Content Provenance and Authenticity's (C2PA) data provenance standard~\cite{rosenthol_2022_c2pa}. CAI's library is the official implementation of C2PA, the leading data provenance standard widely adopted across social media platforms and hardware products. Note, however, that the CRS is not dependent on C2PA and CAI; we simply found it to be the most suitable and adopted data provenance framework to date.

\subsection{Features}

The library has two features: (1) determining whether a single data point considered for inclusion in a dataset would be compliant with the CRS and (2) calculating the overall CRS score of a dataset. We expect feature 1 to be used during the creation of a new dataset, as the dataset author is deciding which data points to include. On the other hand, we expect feature 2 to be used primarily by AI practitioners as they are deciding whether to use a dataset.

\subsubsection{Feature 1}

Feature 1, determining whether a single data point considered for inclusion in a dataset would be compliant with the CRS, requires that data point-level criteria (C2, C3, and C6) be evaluated. The dataset author can pass a considered data point (e.g., an image, video, or audio file) to \textit{DatasetSentinel}. The library will return a boolean indicating whether the data point is compliant. If not, it will list which criteria are violated and provide a description of the reasoning. The schematic overview of this feature is shown in Figure~\ref{fig:use-case-1} (Appendix B). Put into practice, if the dataset author wants their dataset to remain CRS-compliant, they would call this function for every considered data point and drop those for which the assessment is negative.

\subsubsection{Feature 2}

Feature 2, calculating the overall CRS score of a dataset, requires that both dataset- (C1, C4, and C5) and data point-level (C2, C3, and C6) criteria be evaluated. The AI practitioner can provide the full dataset for consideration to \textit{DatasetSentinel}. This dataset can be stored locally or on a dataset sharing platform. The library will return a final CRS score, along with the reasoning: for each criterion, it indicates whether the dataset is compliant, and lists data points that are in violation, if applicable. The schematic overview of this feature is shown in Figure~\ref{fig:use-case-2} (Appendix B). Put into practice, if the AI practitioner wants to ensure the legal and ethical standing of a considered dataset, they would call this function on the dataset, review the assessment, and decide whether it is appropriate to move forward with it.

\subsubsection{Dataset-level Criteria}

Criteria C1, C4, and C5 concern features of the dataset that are determined by the means of distribution. The compliance of a given dataset with these criteria can thus be determined by the inspection of the dataset's page on the distribution platform. For datasets hosted on Hugging Face and Kaggle, \textit{DatasetSentinel} can infer much of this information from the standardized metadata on the dataset's page. For GitHub and custom-hosted datasets, however, there is no standardized way of representing these features, and so \textit{DatasetSentinel} uses an LLM to scan the content of the dataset repository and decide the compliance. To prevent false positive or false negative hits in such cases, \textit{DatasetSentinel} presents the compliance with C1, C4, and C5 for the user to review. The user has the ability to manually override the library's inference.

\subsubsection{Data Point-level Criteria}

Criteria C2, C3, and C6 concern features of data points included in the dataset. A given dataset is compliant with these criteria only when all data points satisfy the criterion. \textit{DatasetSentinel} thus inspects each data point individually and verifies its compliance, which can be derived based on the provenance metadata of the data point (extracted using C2PA) and a set of conditions comparing the provenance information (including the license, whether the content creator opted out of AI training, etc.) to the dataset setting. Unlike dataset-level criteria, these criteria can be clearly determined without user confirmation.

\begin{table}[t]
\centering
\begin{tabular}{p{0.5cm}p{7.2cm}}
\hline
\textbf{\#} & \textbf{Question}                                                                                       \\
\hline
$1$  & How easy is it to navigate the documentation?                                                  \\
$2$  & How understandable is the documentation?                                                       \\
$3$  & How understandable are the tutorials and examples?                                             \\
$4$  & How easily does the library design integrate into your development workflow?                   \\
$5$  & How similar is the structure of the library interface to other libraries you have used before? \\
$6$  & How likely are you to use the library in your workflow while working on a ML project?         
\end{tabular}

\vspace{0.2cm}

\caption{Survey questions used as a part of \textit{DatasetSentinel} library usability evaluation}
\label{tab:questions}
\vspace{0.5cm}
\end{table}

\section{Evaluation}

In this section, we describe two modes of evaluation we employed for \textit{DatasetSentinel} and CRS: an automated code quality assessment and a preliminary user study, surveying $5$ recruited AI experts through a purposive (non-probability) sampling method.

\begin{table}[t]
\centering
\begin{tabular}{lllllllll}
\hline
\textbf{\#}      & \textbf{G} & \textbf{Nat.} & \textbf{Q1}  & \textbf{Q2}  & \textbf{Q3} & \textbf{Q4}  & \textbf{Q5}  & \textbf{Q6}  \\
\hline
P1      & M      & USA       & 7   & 7   & 7  & 4   & 5   & 4   \\
P2      & F      & SWE       & 5   & 5   & 5  & 7   & 5   & 4   \\
P3      & M      & IND       & 5   & 6   & 6  & 3   & 3   & 2   \\
P4      & F      & USA       & 7   & 6   & 6  & 6   & 5   & 7   \\
P5      & M      & USA       & 4   & 5   & 6  & 7   & 3   & 5   \\
P6      & M      & IND       & 6   & 6   & 5  & 4   & 7   & 5   \\
P7      & M      & USA       & 5   & 7   & 6  & 7   & 6   & 6   \\
P8      & M      & SWE       & 7   & 7   & 6  & 7   & 6   & 7   \\
P9      & M      & BGD       & 6   & 5   & 6  & 6   & 7   & 6   \\
P10     & M      & NGA       & 5   & 6   & 5  & 5   & 5   & 6   \\
P11     & M      & USA       & 7   & 7   & 7  & 7   & 7   & 6   \\
P12     & F      & USA       & 7   & 7   & 7  & 7   & 7   & 7   \\
P13     & M      & USA       & 3   & 6   & 6  & 4   & 4   & 4   \\
P14     & M      & AUT       & 6   & 5   & 3  & 3   & 5   & 3   \\
P15     & M      & HKG       & 4   & 3   & 4  & 5   & 6   & 7   \\
\hline
\textbf{Avg.} &        &             & 5.6 & 5.9 & 5.7  & 5.5 & 5.4 & 5.3
\end{tabular}

\vspace{0.2cm}

\caption{Results of the library usability evaluation}
\label{tab:results}
\end{table}

\begin{table*}[t]
\centering
\begin{tabular}{llllllllll}
\hline
\textbf{Dataset} & \textbf{Source} & \textbf{Modality} & \textbf{C1} & \textbf{C2} & \textbf{C3} & \textbf{C4} & \textbf{C5} & \textbf{C6} & \textbf{CRS Score} \\
\hline
SOD4SB & GitHub & Images & \cmark & \cmark & \cmark & \cmark & \xmark & \xmark & C \\
MS COCO & Custom website & Images & \cmark & \xmark & \xmark & \xmark & \xmark & \xmark & F \\
RANDOM People & Hugging Face & Videos & \cmark & \cmark & \cmark & \cmark & \cmark & \xmark & B \\
TikTok Dataset & Kaggle & Videos & \xmark & \xmark & \xmark & \xmark & \xmark & \xmark & G \\
\hline
\end{tabular}
\vspace{0.2cm}

\caption{Results of the CRS case studies on four publicly available datasets. For each dataset, we report whether it satisfies CRS criteria C1 through C6, and to which CRS score this translates.}
\label{tab:results}
\end{table*}

\subsection{Methodology}

\subsubsection{Code Quality}

We used the Wily maintainability score\footnote{\url{https://github.com/tonybaloney/wily}} on the scale from $0$ to $100$, with a higher score reflecting a better evaluation of the complexity, readability, and in-code documentation of the \textit{DatasetSentinel} library. This suite of metrics is based on the Halstead complexity measures~\cite{halstead1977elements}, which have been shown to increase code readability and minimize down-stream fault rates of the evaluated codebase~\cite{khan2023evaluating,coimbra2018correlation}.

\subsubsection{Library Usability}

In addition, to better understand how this prototype would perform in real-life applications, we recruited $14$ participants through a purposive sampling technique to evaluate its usability and robustness. Participants were recruited based on their expertise in the field of AI, and half of them are from the United States (0.5), the rest being from Sweden (0.14), India (0.14), Nigeria (0.07), Bangladesh (0.07) and Austria (0.07). The majority are male (0.85). Participants were asked to implement our script into a database and answer $6$ evaluative questions (presented in Table~\ref{tab:questions}) on a $7$-point Likert scale, where $1$ \= very difficult and $7$ \= very easy, with a high score reflecting greater usability. Participants were given the opportunity to add comments on their experience for a simple qualitative evaluation. The participants were recruited based on their technical expertise in AI and ML. Participants were not recorded; only their written answers were collected and fully anonymized.


\subsection{Results}

\subsubsection{Code Quality}
\textit{DatasetSentinel}'s codebase obtained a mean Wily maintainability score of over $85$, indicating an overall good code quality. The files that were indicated as lower-ranking mostly included connections between our framework and the provenance metadata flags of the C2PA library; we thus make it our priority to keep improving the library in this regard, mainly by adding in-code documentation.

\subsubsection{Library Usability}

The participants' answers are presented in Table~\ref{tab:results}. Quantitatively, we observe an overall positive response to our prototype as all questions are, on average, rated positively ($\geq 5.6$). While these results are preliminary and further work needs to be conducted to further improve the \textit{DatasetSentinel} Library, they are nevertheless encouraging and optimistic. The majority of participants seem to agree that the script is easy to navigate (5.6/7), understand (5.9/7), and well documented with tutorials and examples (5.7/7) (questions \#$1$, \#$2$, and \#$3$). The results relating to ease of integration (question \#$4$, \#$5$, and \#$6$) were slightly less positive (5.5, 5.4, and 5.3 respectively), but encouraging nevertheless.

We would like to highlight that the responses to questions \#$5$ and \#$6$ depend on the type of AI project into which the user is integrating \textit{DatasetSentinel}. For instance, one participant stated that "the primary reason I am unlikely to use this library in my projects is that I almost exclusively work with tabular data" (P3). This is a valid point, although in its current state, our library is designed to address the concerns resulting from image, audio, and video data files. Similarly, another participant stated that "my projects aren't really about ethics, which is the only reason I put only a 4" (P1). We find this statement to be a good reflection on the general dissociation found among practitioners between AI applications and ethics. Another participant stated that the CRS score's function was unclear, as they could not find any information online regarding this tool and asked for clarifications: "It is unclear whether the CRS score is something you invented or an agreed-upon standard. Searching for the CRS score take me to the Canadian government site..." (P2). This confusion was caused by the fact that we could not reveal the manuscript where we introduced and explained the CRS score to maintain high discretion and total anonymity. Similarly, another participant stated that "I wanted to learn more about C2PA – a brief explanation and link would be great" (P13). We agree with this comment, as we believe it is crucial not to assume that every AI practitioner might be familiarised with C2PA, and how does the CRS score differ from it: "What's the difference between [DatasetSentinel] library and the C2PA Python library? Is C2PA more low-level and this one provides nicer abstractions, or is there something functionally different?" (P13). We have addressed these comments and updated the documentation. 
Other participants seem to appreciate the value of this work, as one mentioned that "I think that this library is very well organized, thoughtful and important for today's modern tech world" (P4) and another that "this project looks incredibly useful and helpful" (P5).

\section{Case Studies}

To put the CRS framework and \textit{DatasetSentinel} library to practice, we applied them to four open-source datasets from different modes of distribution (GitHub, Hugging Face, Kaggle, and custom website). Next, we briefly describe these datasets and present their CRS assessment. The results are summarized in Table~\ref{tab:results}.

\subsection{Use Case: SOD4SB}

The SOD4SB dataset~\cite{kondo2023mva2023}, released as a part of the MVA2023 Spotting Birds challenge, contains $39,070$ images annotated with bounding boxes of birds. These images were taken by the dataset's authors. The dataset is distributed through GitHub. As with the previous dataset, it is not compliant with criterion C6. Additionally, it is not compliant with criterion C5, as there is no trace log of changes. This results in the CRS Score "C".

\subsection{Use Case: MS COCO}

The MS COCO dataset~\cite{lin2014microsoft} contains over $300,000$ images with annotations for object detection, segmentation, captioning, and keypoint detection. The images were gathered from Flickr. The dataset is distributed through a custom website. As with the previous dataset, it is not compliant with the criteria C5 and C6. Additionally, it is not compliant with criterion C4, as there is no opting-out mechanism; C3, as the data points with inconclusive provenance metadata are not flagged; and C2, as some data points are used against their license. This results in the CRS Score "F".

\begin{figure*}[t]
  \centering
  \includegraphics[width=0.84\textwidth]{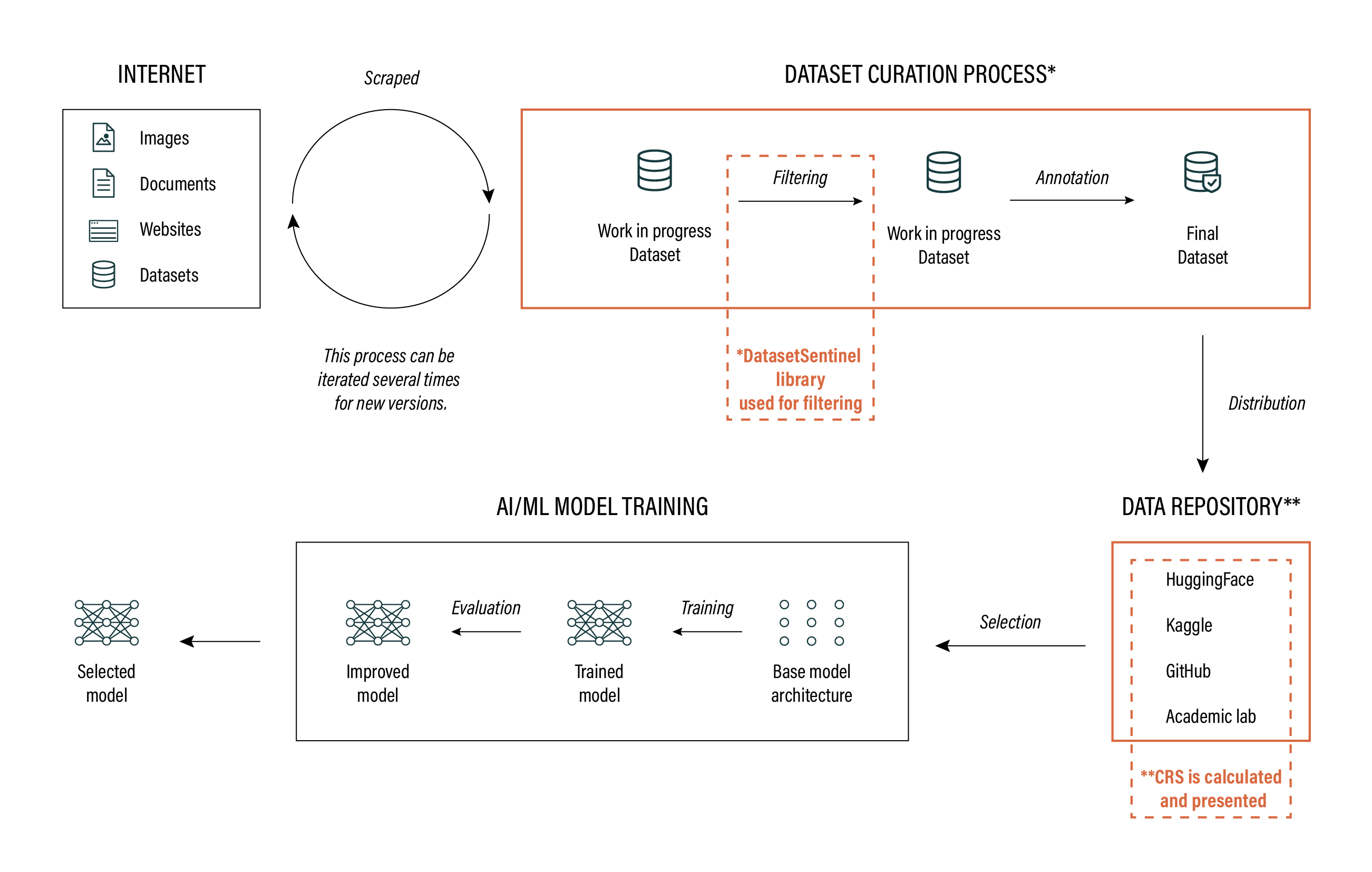}
  
  \caption{A schematic overview of the AI ecosystem workflow with the main stages of dataset and model development}
  \label{fig:ai-worfklow}
\end{figure*}

\subsection{Use Case: RANDOM People}

The RANDOM People dataset\footnote{\url{https://anonymous.4open.science/r/random-people-dataset-D70F/}} contains videos with human protagonists performing actions around the house, generated using a pose-transfer AI model, along with the annotations of these actions. The identities used as a reference for pose transfer were consenting individuals gathered by the dataset authors, and the driving videos were from an open-source database whose creator had permission from all depicted participants. The dataset is distributed on Hugging Face. It is not compliant with criterion C6, as the dataset source and the retention period are not added to the provenance metadata of data points, resulting in the CRS Score "B".



\subsection{Use Case: The TikTok Dataset}

The TikTok Dataset~\cite{jafarian2021learning} contains $300$ dance videos, $10$ to $15$ seconds in length, sourced from TikTok. Additional 3D representations are also provided. The dataset is distributed through Kaggle. As with the previous dataset, it is not compliant with criteria C2, C3, C4, C5, and C6. Additionally, it is not compliant with criterion C1, as the sourcing, filtering, and pre-processing are not detailed at a level that would enable reproducing the dataset. This results in the CRS Score "G".

\section{Discussion}

By proposing a set of four practical principles to consider for dataset compliance in the context of AI, we aim to provide a framework that raises the discussion on the legality and ethics of AI applications. However, similar to most principles, these can be interpreted as highly conceptual and disconnected from current practices, often making them either irrelevant or challenging to implement. Precisely for this reason, we attempted to move away from a purely descriptive contribution to the literature, and provide a tangible and prescriptive approach through our CRS tool and \emph{DatasetSentinel} library. We highlight the specific points of the AI workflow at which we target our contribution, aiming to reduce the misuse of personal data for GAI training models and applications by introducing traceability and accountability of the datasets used for harmful purposes. 

To this end, the first line of defense is with the \emph{DatasetSentinel} library, which can be used by practitioners to filter the collected data. Using provenance metadata, the tool ensures that the data is compliant with the purpose of the dataset. The second line of defense is the CRS score, which calculates and informs the practitioners about the dataset’s compliance with the practical principles embedded in its structure. These two intervention points in the life cycle of a dataset are illustrated in Figure~\ref{fig:ai-worfklow}. 

We believe that the benefits of implementing this tool are twofold. In the long term, it benefits the AI field and, more broadly, society as a whole. Over time, poorly rated datasets (E and below) would stop being used as much and eventually become less impactful. We ground this belief in studies about consumers' quality standards expectations, showing that 92\% of consumers tend to purchase products with at least a $4$-star rating\footnote{\url{https://explodingtopics.com/blog/online-review-stats}}. We believe the field of AI is no different. To induce this effect in AI practitioners while choosing datasets, we propose accompanying visuals for the CRS scores shown in Appendix~\ref{appendix:crs-scale}. In the short term, it benefits the individual user as it removes the heavy lifting of manually conducting this type of analysis and helps protect themselves from any liability of data misuse. 

We intend to render it more challenging for defendants accused of malicious activity through AI applications to plead ignorance about the nature or compliance of any given dataset. In the eventuality of a legal demand, the CRS score enables developers and regulators to gauge and verify the transparency, accountability, and security of any given dataset, with the ultimate objective of providing traceability and accountability. By doing so, we hope to help reduce the gap between digital technological innovation and ethics by providing a framework to responsibility, liability, and legal enforcement of data malpractices in the context of AI.

In the future, dataset-sharing platforms may adopt this tool on their end, which would remove the heavy lifting (of running this analysis) from individual users. Shown in Figures~\ref{fig:mockup-hf}, \ref{fig:mockup-github}, and \ref{fig:mockup-lab-website} (Appendix C) are mockups that fictitiously contextualizes the CRS score in an online repository, as practitioners would perceive it. As observed in these mock-ups, the CRS score seemingly integrates with the rest of the dataset's information, while providing a clear reading .

Regarding the adoption of CRS and the \textit{DatasetSentinel} library, we do not expect them to be a mandatory requirement but rather a tool to support the AI community. By providing an overview of the compliance of any given dataset, both dataset owners and users can better reflect on their responsibility and liability towards the AI community, and make a more informed decision on the resources they use in their projects.

We are witnessing a growing interest among software and hardware companies in tracing the provenance of media in an attempt to fight misinformation and other malicious content. This trend is manifesting itself, for example, by an uptick of organizations joining coalitions such as the Coalition for Content Provenance and Authenticity (C2PA) \cite{rosenthol_2022_c2pa}. It seems that there is a growing trend towards data traceability and immutability within the digital sphere. We therefore reiterate our belief in this project and its potential positive impact within the field of AI.

\section{Limitations}

As our prototype is in its infancy, we acknowledge its limitations and that there is still much research to be conducted until this framework can become a standard for ethical GAI use. For instance, as data provenance technologies are just rolling out, the majority of digital media available online still lacks provenance metadata. Nonetheless, many technological companies – both in software and hardware – are starting to deploy or announce the integration of data provenance technologies into their products. There are indications that this trend is becoming more and more common, as users express concerns over the use of their images, artwork and intellectual property; and companies are attempting to solve this. For instance, we do not discard the possibility of smartphone operating systems introducing an "opt-out" feature for all (or only selected images and videos) taken on the smartphone for AI training. As such, we expect that, within a few years, the vast majority of new digital media distributed on the internet will have provenance metadata. Another limitation is that the library is dependent on the existing data provenance protocols. To that end, our library can only analyze data types that are supported by these protocols and other dependencies. This should not pose a problem for most current use cases, as the protocols support the most common data types for image, video, audio, and 3D objects. Still, moving forward, this dependency could introduce a delay in introducing support for new data types. 


\section{Conclusion}

We call for a larger discussion confronting the unsustainable dataset practices in the AI community. While we recognize that the dataset sharing platforms have substantial power to influence the practical rules and guidelines, we argue that a value shift is also needed. Specifically, a broader awareness and appreciation of ethical and legal considerations surrounding datasets must be established for the rules and guidelines of dataset sharing platforms to have a meaningful impact. Our framework and tangible outputs can serve as a springboard for piloting and implementing these values into existing workflows.


\newpage

\bibliographystyle{ACM-Reference-Format}
\bibliography{sample-base}

@String{Computing = "Computing" }

@String{Computer = "{IEEE} Computer" }

@String{Academic = "Academic Press" }

@String{Springer = "Springer-Verlag" }

@article{birhane_2021_multimodal,
  title={Multimodal datasets: misogyny, pornography, and malignant stereotypes},
  author={Birhane, Abeba and Prabhu, Vinay Uday and Kahembwe, Emmanuel},
  journal={arXiv preprint arXiv:2110.01963},
  year={2021}
}

@article{schuhmann_2022_laion,
  title={{LAION-5B}: An open large-scale dataset for training next generation image-text models},
  author={Schuhmann, Christoph and Beaumont, Romain and Vencu, Richard and Gordon, Cade and Wightman, Ross and Cherti, Mehdi and Coombes, Theo and Katta, Aarush and Mullis, Clayton and Wortsman, Mitchell and others},
  journal={Advances in Neural Information Processing Systems},
  volume={35},
  pages={25278--25294},
  year={2022}
}

@article{oppenlaender_2023_text,
  title={Text-to-Image Generation: Perceptions and Realities},
  author={Oppenlaender, Jonas and Visuri, Aku and Paananen, Ville and Linder, Rhema and Silvennoinen, Johanna},
  journal={arXiv preprint arXiv:2303.13530},
  year={2023}
}

@inproceedings{schramowski_2023_safe,
  title={Safe latent diffusion: Mitigating inappropriate degeneration in diffusion models},
  author={Schramowski, Patrick and Brack, Manuel and Deiseroth, Bj{\"o}rn and Kersting, Kristian},
  booktitle={Proceedings of the IEEE/CVF Conference on Computer Vision and Pattern Recognition},
  pages={22522--22531},
  year={2023}
}

@article{gonzalez_2020_state,
  title={The State of the ML-universe: 10 Years of Artificial Intelligence \& Machine Learning Software Development on GitHub},
  author={Danielle Gonzalez and Thomas Zimmermann and Nachiappan Nagappan},
  journal={2020 IEEE/ACM 17th International Conference on Mining Software Repositories (MSR)},
  year={2020},
  pages={431-442},
  url={https://api.semanticscholar.org/CorpusID:220117963}
}

@article{rosenthol_2020_content,
  title={The content authenticity initiative: Setting the standard for digital content attribution},
  author={Rosenthol, Leonard and Parsons, Andy and Scouten, Eric and Aythora, Jatin and MacCormack, Bruce and England, Paul and Levallee, Marc and Dotan, Jonathan and Hanna, Sherif and Farid, Hany and others},
  journal={Adobe Whitepaper},
  year={2020}
}

@inproceedings{rosenthol_2022_c2pa,
  title={C2PA: the world’s first industry standard for content provenance},
  author={Rosenthol, Leonard},
  booktitle={Applications of Digital Image Processing XLV},
  volume={12226},
  pages={122260P},
  year={2022},
  organization={SPIE}
}

@inproceedings{Paszke_2019_PyTorch,
  title={PyTorch: An Imperative Style, High-Performance Deep Learning Library},
  author={Adam Paszke and Sam Gross and Francisco Massa and Adam Lerer and James Bradbury and Gregory Chanan and Trevor Killeen and Zeming Lin and Natalia Gimelshein and Luca Antiga and Alban Desmaison and Andreas K{\"o}pf and Edward Yang and Zach DeVito and Martin Raison and Alykhan Tejani and Sasank Chilamkurthy and Benoit Steiner and Lu Fang and Junjie Bai and Soumith Chintala},
  booktitle={Neural Information Processing Systems},
  year={2019},
  url={https://api.semanticscholar.org/CorpusID:202786778}
}

@inproceedings{Abadi_2016_TensorFlow,
  title={TensorFlow: A system for large-scale machine learning},
  author={Mart{\'i}n Abadi and Paul Barham and (...) and Xiaoqiang Zhang},
  booktitle={USENIX Symposium on Operating Systems Design and Implementation},
  year={2016},
  url={https://api.semanticscholar.org/CorpusID:6287870}
}

@software{hannun_2023_mlx,
  author = {Awni Hannun and Jagrit Digani and Angelos Katharopoulos and Ronan Collobert},
  title = {{MLX}: Efficient and flexible machine learning on Apple silicon},
  url = {https://github.com/ml-explore},
  version = {0.0},
  year = {2023},
}

@article{wolf_2019_HuggingFace,
  title={HuggingFace's Transformers: State-of-the-art Natural Language Processing},
  author={Thomas Wolf and Lysandre Debut and Victor Sanh and Julien Chaumond and Clement Delangue and Anthony Moi and Pierric Cistac and Tim Rault and R{\'e}mi Louf and Morgan Funtowicz and Jamie Brew},
  journal={ArXiv},
  year={2019},
  volume={abs/1910.03771},
  url={https://api.semanticscholar.org/CorpusID:208117506}
}

@article{zhang_2023_navigating,
  title={Navigating Privacy and Copyright Challenges Across the Data Lifecycle of Generative AI},
  author={Dawen Zhang and Boming Xia and Yue Liu and Xiwei Xu and Thong Hoang and Zhenchang Xing and Mark Staples and Qinghua Lu and Liming Zhu},
  journal={ArXiv},
  year={2023},
  volume={abs/2311.18252},
  url={https://api.semanticscholar.org/CorpusID:265506320}
}

@article{polyzotis_2018_data,
  title={Data lifecycle challenges in production machine learning: a survey},
  author={Polyzotis, Neoklis and Roy, Sudip and Whang, Steven Euijong and Zinkevich, Martin},
  journal={ACM SIGMOD Record},
  volume={47},
  number={2},
  pages={17--28},
  year={2018},
  publisher={ACM New York, NY, USA}
}

@article{paullada_2020_data,
  title={Data and its (dis)contents: A survey of dataset development and use in machine learning research},
  author={Amandalynne Paullada and Inioluwa Deborah Raji and Emily M. Bender and Emily L. Denton and A. Hanna},
  journal={Patterns},
  year={2020},
  volume={2},
  url={https://api.semanticscholar.org/CorpusID:228084012}
}

@article{chugunkov_2018_creation,
  title={Creation of datasets from open sources},
  author={Ilya V. Chugunkov and Dmitry V. Kabak and Viktor N. Vyunnikov and Roman E. Aslanov},
  journal={2018 IEEE Conference of Russian Young Researchers in Electrical and Electronic Engineering (EIConRus)},
  year={2018},
  pages={295-297},
  url={https://api.semanticscholar.org/CorpusID:3899704}
}

@article{koch_2021_reduced,
  title={Reduced, Reused and Recycled: The Life of a Dataset in Machine Learning Research},
  author={Bernard Koch and Emily L. Denton and A. Hanna and Jacob Gates Foster},
  journal={ArXiv},
  year={2021},
  volume={abs/2112.01716},
  url={https://api.semanticscholar.org/CorpusID:244894836}
}

@article{scheuerman_2021_DoDH,
  title={Do Datasets Have Politics? Disciplinary Values in Computer Vision Dataset Development},
  author={Morgan Klaus Scheuerman and Emily L. Denton and A. Hanna},
  journal={Proceedings of the ACM on Human-Computer Interaction},
  year={2021},
  volume={5},
  pages={1 - 37},
  url={https://api.semanticscholar.org/CorpusID:236965639}
}

@article{villalobos_2022_WillWR,
  title={Will we run out of data? An analysis of the limits of scaling datasets in Machine Learning},
  author={Pablo Villalobos and Jaime Sevilla and Lennart Heim and Tamay Besiroglu and Marius Hobbhahn and An Chang Ho},
  journal={ArXiv},
  year={2022},
  volume={abs/2211.04325},
  url={https://api.semanticscholar.org/CorpusID:253397775}
}

@article{halevy_2009_TheUE,
  title={The Unreasonable Effectiveness of Data},
  author={Alon Y. Halevy and Peter Norvig and Fernando C Pereira},
  journal={IEEE Intelligent Systems},
  year={2009},
  volume={24},
  pages={8-12},
  url={https://api.semanticscholar.org/CorpusID:14300215}
}

@article{chiang_2022_ExploringTE,
  title={Exploring the Effects of Machine Learning Literacy Interventions on Laypeople’s Reliance on Machine Learning Models},
  author={Chun-Wei Chiang and Ming Yin},
  journal={27th International Conference on Intelligent User Interfaces},
  year={2022},
  url={https://api.semanticscholar.org/CorpusID:247585115}
}

@article{paullada_2020_DataAI,
  title={Data and its (dis)contents: A survey of dataset development and use in machine learning research},
  author={Amandalynne Paullada and Inioluwa Deborah Raji and Emily M. Bender and Emily L. Denton and A. Hanna},
  journal={Patterns},
  year={2020},
  volume={2},
  url={https://api.semanticscholar.org/CorpusID:228084012}
}

@article{khder_2021_WebSO,
  title={Web Scraping or Web Crawling: State of Art, Techniques, Approaches and Application},
  author={Moaiad Ahmad Khder},
  journal={International Journal of Advances in Soft Computing and its Applications},
  year={2021},
  url={https://api.semanticscholar.org/CorpusID:245584401}
}

@article{cong_2021_DataPI,
  title={Data pricing in machine learning pipelines},
  author={Zicun Cong and Xuan Luo and Jian Pei and Feida Zhu and Yong Zhang},
  journal={Knowledge and Information Systems},
  year={2021},
  volume={64},
  pages={1417 - 1455},
  url={https://api.semanticscholar.org/CorpusID:237194666}
}

@article{hutchinson_2020_TowardsAF,
  title={Towards Accountability for Machine Learning Datasets: Practices from Software Engineering and Infrastructure},
  author={Ben Hutchinson and Andrew Smart and A. Hanna and Emily L. Denton and Christina Greer and Oddur Kjartansson and Parker Barnes and Margaret Mitchell},
  journal={Proceedings of the 2021 ACM Conference on Fairness, Accountability, and Transparency},
  year={2020},
  url={https://api.semanticscholar.org/CorpusID:225067460}
}

@article{sambasivan_2021_EveryoneWT,
  title={“Everyone wants to do the model work, not the data work”: Data Cascades in High-Stakes AI},
  author={Nithya Sambasivan and Shivani Kapania and Hannah Highfill and Diana Akrong and Praveen K. Paritosh and Lora Aroyo},
  journal={Proceedings of the 2021 CHI Conference on Human Factors in Computing Systems},
  year={2021},
  url={https://api.semanticscholar.org/CorpusID:231829607}
}

@article{heger_2022_UnderstandingML,
  title={Understanding Machine Learning Practitioners' Data Documentation Perceptions, Needs, Challenges, and Desiderata},
  author={Amy Kathleen Heger and Elizabeth B. Marquis and Mihaela Vorvoreanu and Hanna M. Wallach and Jenn Wortman Vaughan},
  journal={Proceedings of the ACM on Human-Computer Interaction},
  year={2022},
  volume={6},
  pages={1 - 29},
  url={https://api.semanticscholar.org/CorpusID:249431472}
}

@article{banh_2023_GenerativeAI,
  title={Generative artificial intelligence},
  author={Leonardo Banh and Gero Strobel},
  journal={Electronic Markets},
  year={2023},
  volume={33},
  pages={1-17},
  url={https://api.semanticscholar.org/CorpusID:265675536}
}

@article{brown_2020_LanguageMA,
  title={Language Models are Few-Shot Learners},
  author={Tom B. Brown and Benjamin Mann and (...) and Dario Amodei},
  journal={ArXiv},
  year={2020},
  volume={abs/2005.14165},
  url={https://api.semanticscholar.org/CorpusID:218971783}
}

@article{khan2023evaluating,
  title={Evaluating the effectiveness of decomposed Halstead Metrics in software fault prediction},
  author={Khan, Bilal and Nadeem, Aamer},
  journal={PeerJ Computer Science},
  volume={9},
  pages={e1647},
  year={2023},
  publisher={PeerJ Inc.}
}

@inproceedings{coimbra2018correlation,
  title={A correlation analysis between halstead complexity measures and other software measures},
  author={Coimbra, Rodrigo Tavares and Resende, Ant{\^o}nio and Terra, Ricardo},
  booktitle={2018 XLIV Latin American Computer Conference (CLEI)},
  pages={31--39},
  year={2018},
  organization={IEEE}
}

@article{touvron_2023_LLaMAOA,
  title={LLaMA: Open and Efficient Foundation Language Models},
  author={Hugo Touvron and Thibaut Lavril and Gautier Izacard and Xavier Martinet and Marie-Anne Lachaux and Timoth{\'e}e Lacroix and Baptiste Rozi{\`e}re and Naman Goyal and Eric Hambro and Faisal Azhar and Aurelien Rodriguez and Armand Joulin and Edouard Grave and Guillaume Lample},
  journal={ArXiv},
  year={2023},
  volume={abs/2302.13971},
  url={https://api.semanticscholar.org/CorpusID:257219404}
}

@article{zhang_2023_TexttoimageDM,
  title={Text-to-image Diffusion Models in Generative AI: A Survey},
  author={Chenshuang Zhang and Chaoning Zhang and Mengchun Zhang and In-So Kweon},
  journal={ArXiv},
  year={2023},
  volume={abs/2303.07909},
  url={https://api.semanticscholar.org/CorpusID:257505012}
}

@article{bie_2023_RenAIssanceAS,
  title={RenAIssance: A Survey into AI Text-to-Image Generation in the Era of Large Model},
  author={Fengxiang Bie and Yibo Yang and Zhongzhu Zhou and Adam Ghanem and Minjia Zhang and Zhewei Yao and Xiaoxia Wu and Connor Holmes and Pareesa Ameneh Golnari and David A. Clifton and Yuxiong He and Dacheng Tao and Shuaiwen Leon Song},
  journal={ArXiv},
  year={2023},
  volume={abs/2309.00810},
  url={https://api.semanticscholar.org/CorpusID:265821110}
}

@article{wang_2023_VideoComposerCV,
  title={VideoComposer: Compositional Video Synthesis with Motion Controllability},
  author={Xiang Wang and Hangjie Yuan and Shiwei Zhang and Dayou Chen and Jiuniu Wang and Yingya Zhang and Yujun Shen and Deli Zhao and Jingren Zhou},
  journal={ArXiv},
  year={2023},
  volume={abs/2306.02018},
  url={https://api.semanticscholar.org/CorpusID:259075720}
}

@article{wu_2022_TuneAVideoOT,
  title={Tune-A-Video: One-Shot Tuning of Image Diffusion Models for Text-to-Video Generation},
  author={Jay Zhangjie Wu and Yixiao Ge and Xintao Wang and Weixian Lei and Yuchao Gu and Wynne Hsu and Ying Shan and Xiaohu Qie and Mike Zheng Shou},
  journal={2023 IEEE/CVF International Conference on Computer Vision (ICCV)},
  year={2022},
  pages={7589-7599},
  url={https://api.semanticscholar.org/CorpusID:254974187}
}

@article{khachatryan_2023_Text2VideoZeroTD,
  title={Text2Video-Zero: Text-to-Image Diffusion Models are Zero-Shot Video Generators},
  author={Levon Khachatryan and Andranik Movsisyan and Vahram Tadevosyan and Roberto Henschel and Zhangyang Wang and Shant Navasardyan and Humphrey Shi},
  journal={2023 IEEE/CVF International Conference on Computer Vision (ICCV)},
  year={2023},
  pages={15908-15918},
  url={https://api.semanticscholar.org/CorpusID:257687280}
}

@article{singer_2022_MakeAVideoTG,
  title={Make-A-Video: Text-to-Video Generation without Text-Video Data},
  author={Uriel Singer and Adam Polyak and Thomas Hayes and Xiaoyue Yin and Jie An and Songyang Zhang and Qiyuan Hu and Harry Yang and Oron Ashual and Oran Gafni and Devi Parikh and Sonal Gupta and Yaniv Taigman},
  journal={ArXiv},
  year={2022},
  volume={abs/2209.14792},
  url={https://api.semanticscholar.org/CorpusID:252595919}
}

@article{natsiou_2021_AudioRF,
  title={Audio representations for deep learning in sound synthesis: A review},
  author={Anastasia Natsiou and Se{\'a}n O'Leary},
  journal={2021 IEEE/ACS 18th International Conference on Computer Systems and Applications (AICCSA)},
  year={2021},
  pages={1-8},
  url={https://api.semanticscholar.org/CorpusID:245827795}
}

@misc{thiel2023generative,
  title={Generative ML and CSAM: Implications and Mitigations},
  author={Thiel, David and Stroebel, Melissa and Portnoff, Rebecca},
  year={2023},
  publisher={Stanford Digital Repository. https://doi. org/10.25740/jv206yg3793}
}

@inproceedings{bhardwaj2024machine,
  title={Machine learning data practices through a data curation lens: An evaluation framework},
  author={Bhardwaj, Eshta and Gujral, Harshit and Wu, Siyi and Zogheib, Ciara and Maharaj, Tegan and Becker, Christoph},
  booktitle={The 2024 ACM Conference on Fairness, Accountability, and Transparency},
  pages={1055--1067},
  year={2024}
}

@inproceedings{ajmani2024data,
  title={Data Agency Theory: A Precise Theory of Justice for AI Applications},
  author={Ajmani, Leah and Stapleton, Logan and Houtti, Mo and Chancellor, Stevie},
  booktitle={The 2024 ACM Conference on Fairness, Accountability, and Transparency},
  pages={631--641},
  year={2024}
}

@inproceedings{kalchbrenner_2018_EfficientNA,
  title={Efficient Neural Audio Synthesis},
  author={Nal Kalchbrenner and Erich Elsen and Karen Simonyan and Seb Noury and Norman Casagrande and Edward Lockhart and Florian Stimberg and A{\"a}ron van den Oord and Sander Dieleman and Koray Kavukcuoglu},
  booktitle={International Conference on Machine Learning},
  year={2018},
  url={https://api.semanticscholar.org/CorpusID:3524525}
}

@article{xue_2024_AuffusionLT,
  title={Auffusion: Leveraging the Power of Diffusion and Large Language Models for Text-to-Audio Generation},
  author={Jinlong Xue and Yayue Deng and Yingming Gao and Ya Li},
  journal={ArXiv},
  year={2024},
  volume={abs/2401.01044},
  url={https://api.semanticscholar.org/CorpusID:266725678}
}

@article{zhao_2023_ASO,
  title={A Survey of Large Language Models},
  author={Wayne Xin Zhao and Kun Zhou and Junyi Li (...) and Ji-rong Wen},
  journal={ArXiv},
  year={2023},
  volume={abs/2303.18223},
  url={https://api.semanticscholar.org/CorpusID:257900969}
}

@article{singh_2023_ASO,
  title={A Survey of AI Text-to-Image and AI Text-to-Video Generators},
  author={Aditi Singh},
  journal={2023 4th International Conference on Artificial Intelligence, Robotics and Control (AIRC)},
  year={2023},
  pages={32-36},
  url={https://api.semanticscholar.org/CorpusID:264977095}
}

@article{liu_2021_AIEmpoweredPV,
  title={AI-Empowered Persuasive Video Generation: A Survey},
  author={Chang Liu and Han Yu},
  journal={ACM Computing Surveys},
  year={2021},
  volume={55},
  pages={1 - 31},
  url={https://api.semanticscholar.org/CorpusID:245329411}
}

@article{zhang_2023_ASO,
  title={A Survey on Audio Diffusion Models: Text To Speech Synthesis and Enhancement in Generative AI},
  author={Chenshuang Zhang and Chaoning Zhang and Sheng Zheng and Mengchun Zhang and Maryam Qamar and Sung-Ho Bae and In-So Kweon},
  journal={ArXiv},
  year={2023},
  volume={abs/2303.13336},
  url={https://api.semanticscholar.org/CorpusID:257913174}
}

@article{maslej_2023_ArtificialII,
  title={Artificial Intelligence Index Report 2023},
  author={Nestor Maslej and Loredana Fattorini and Erik Brynjolfsson and John Etchemendy and Katrina Ligett and Terah Lyons and James Manyika and Helen Ngo and Juan Carlos Niebles and Vanessa Parli and Yoav Shoham and Russell Wald and Jack Clark and Ray Perrault},
  journal={ArXiv},
  year={2023},
  volume={abs/2310.03715}
}

@article{wei_2022_EmergentAO,
  title={Emergent Abilities of Large Language Models},
  author={Jason Wei and Yi Tay and Rishi Bommasani and Colin Raffel and Barret Zoph and Sebastian Borgeaud and Dani Yogatama and Maarten Bosma and Denny Zhou and Donald Metzler and Ed Huai-hsin Chi and Tatsunori Hashimoto and Oriol Vinyals and Percy Liang and Jeff Dean and William Fedus},
  journal={Trans. Mach. Learn. Res.},
  year={2022},
  volume={2022}
}

@article{valmeekam_2023_OnTP,
  title={On the Planning Abilities of Large Language Models (A Critical Investigation with a Proposed Benchmark)},
  author={Karthik Valmeekam and Sarath Sreedharan and Matthew Marquez and Alberto Olmo Hernandez and Subbarao Kambhampati},
  journal={ArXiv},
  year={2023},
  volume={abs/2302.06706}
}

@inproceedings{bohacek_2023_TheUA,
  title={The Unseen A+ Student: Navigating the Impact of Large Language Models in the Classroom},
  author={Matyas Bohacek},
  year={2023},
  url={https://api.semanticscholar.org/CorpusID:262939886}
}

@article{casal_2023_CanLD,
  title={Can linguists distinguish between ChatGPT/AI and human writing?: A study of research ethics and academic publishing},
  author={J. Elliott Casal and Matthew Kessler},
  journal={Research Methods in Applied Linguistics},
  year={2023},
  url={https://api.semanticscholar.org/CorpusID:260713371}
}

@article{goodfellow_2014_GenerativeAN,
  title={Generative adversarial networks},
  author={Ian J. Goodfellow and Jean Pouget-Abadie and Mehdi Mirza and Bing Xu and David Warde-Farley and Sherjil Ozair and Aaron C. Courville and Yoshua Bengio},
  journal={Communications of the ACM},
  year={2014},
  volume={63},
  pages={139 - 144},
  url={https://api.semanticscholar.org/CorpusID:1033682}
}

@article{karras_2018_ASG,
  title={A Style-Based Generator Architecture for Generative Adversarial Networks},
  author={Tero Karras and Samuli Laine and Timo Aila},
  journal={2019 IEEE/CVF Conference on Computer Vision and Pattern Recognition (CVPR)},
  year={2018},
  pages={4396-4405},
  url={https://api.semanticscholar.org/CorpusID:54482423}
}

@article{nightingale_2022_AIsynthesizedFA,
  title={AI-synthesized faces are indistinguishable from real faces and more trustworthy},
  author={Sophie J. Nightingale and Hany Farid},
  journal={Proceedings of the National Academy of Sciences of the United States of America},
  year={2022},
  volume={119},
  url={https://api.semanticscholar.org/CorpusID:246827447}
}

@article{pan_2023_DataPI,
  title={Data Provenance in Security and Privacy},
  author={Bofeng Pan and Natalia Stakhanova and Suprio Ray},
  journal={ACM Computing Surveys},
  year={2023},
  volume={55},
  pages={1 - 35},
  url={https://api.semanticscholar.org/CorpusID:258259369}
}

@inproceedings{chen_2017_DataPA,
  title={Data Provenance at Internet Scale: Architecture, Experiences, and the Road Ahead},
  author={Ang Chen and Yang Wu and Andreas Haeberlen and Boon Thau Loo and Wenchao Zhou},
  booktitle={Conference on Innovative Data Systems Research},
  year={2017},
  url={https://api.semanticscholar.org/CorpusID:559852}
}

@incollection{magagna_2020_data,
  title={Data provenance},
  author={Magagna, Barbara and Goldfarb, Doron and Martin, Paul and Atkinson, Malcolm and Koulouzis, Spiros and Zhao, Zhiming},
  booktitle={Towards Interoperable Research Infrastructures for Environmental and Earth Sciences: A Reference Model Guided Approach for Common Challenges},
  pages={208--225},
  year={2020},
  publisher={Springer}
}

@inproceedings{engram_2021_ProactivePP,
  title={Proactive Provenance Policies for Automatic Cryptographic Data Centric Security},
  author={Shamaria Engram and Tyler Kaczmarek and Alice Lee and David Bigelow},
  booktitle={International Provenance and Annotation Workshop},
  year={2021},
  url={https://api.semanticscholar.org/CorpusID:235266153}
}

@book{halstead1977elements,
  title={Elements of Software Science (Operating and programming systems series)},
  author={Halstead, Maurice H},
  year={1977},
  publisher={Elsevier Science Inc.}
}

@inproceedings{sidnam_2022_UsableCP,
  title={Usable Cryptographic Provenance: A Proactive Complement to Fact-Checking for Mitigating Misinformation},
  author={Emily Sidnam-Mauch and Bernat Ivancsics and Ayana Monroe and Evelien Bergrath Washington and Errol Francis and Kelly E. Caine and Joseph Bonneau and Susan E. McGregor},
  booktitle={ICWSM Workshops},
  year={2022},
  url={https://api.semanticscholar.org/CorpusID:249668756}
}

@article{werder_2022_EstablishingDP,
  title={Establishing Data Provenance for Responsible Artificial Intelligence Systems},
  author={Karl Werder and Balasubramaniam Ramesh and Rongen Zhang},
  journal={ACM Transactions on Management Information Systems (TMIS)},
  year={2022},
  volume={13},
  pages={1 - 23},
  url={https://api.semanticscholar.org/CorpusID:247395133}
}

@article{cofone2021beyond,
  title={Beyond data ownership},
  author={Cofone, Ignacio},
  journal={Cardozo L. Rev.},
  volume={43},
  pages={501},
  year={2021},
  publisher={HeinOnline}
}

@misc{noauthor_general_2018,
	title = {General Data Protection Regulation ({GDPR}) – Official Legal Text},
    author = {},
	url = {https://gdpr-info.eu/},
	shorttitle = {{GDPR}},
	titleaddon = {General Data Protection Regulation ({GDPR})},
	urldate = {2021-06-18},
	date = {2018},
    year = {2018}
}

@article{hummel_own_2020,
	title = {Own Data? Ethical Reflections on Data Ownership},
	url = {https://www.researchgate.net/publication/342188494_Own_Data_Ethical_Reflections_on_Data_Ownership},
	doi = {10.1007/s13347-020-00404-9},
	journaltitle = {Springer},
	author = {Hummel, Patrik and Braun, Matthias and Dabrock, Peter},
	date = {2020-06-15},
}

@online{noauthor_california_2018,
	title = {California Consumer Privacy Act 2018},
	url = {https://oag.ca.gov/privacy/ccpa},
	shorttitle = {{CCPA}},
	urldate = {2023-07-04},
	date = {2018},
    year = {2018}
}

@article{adjerid_choice_2019,
	title = {Choice Architecture, Framing, and Cascaded Privacy Choices},
	volume = {65},
	url = {https://pubsonline.informs.org/doi/pdf/10.1287/mnsc.2018.3028},
	pages = {2267--2290},
	number = {5},
	journaltitle = {Management Science},
	author = {Adjerid, Idris and Acquisti, Alessandro and Loewenstein, George},
	date = {2019-05-05},
}

@article{asswad_data_2021,
	title = {Data Ownership: A Survey},
	volume = {12},
	url = {https://www.mdpi.com/2078-2489/12/11/465},
	number = {465},
	journaltitle = {Information},
	author = {Asswad, Jad and Marx Gómez, Jorge},
	date = {2021-11-10},
}

@article{draper_corporate_2019,
	title = {The corporate cultivation of digital resignation},
	volume = {21},
	issn = {1461-4448},
	url = {https://doi.org/10.1177/1461444819833331},
	doi = {10.1177/1461444819833331},
	pages = {1824--1839},
	number = {8},
	journaltitle = {New Media \& Society},
	author = {Draper, Nora A and Turow, Joseph},
	urldate = {2023-12-09},
	date = {2019-08-01},
	langid = {english},
	note = {Publisher: {SAGE} Publications},
}

@article{solove_introduction_2012,
	title = {Introduction: Privacy Self-Management and the Consent Dilemma Symposium: Privacy and Technology},
	volume = {126},
	url = {https://heinonline.org/HOL/P?h=hein.journals/hlr126&i=1910},
	shorttitle = {Introduction},
	pages = {1880--1903},
	number = {7},
	journal = {Harvard Law Review},
	author = {Solove, Daniel J.},
	urldate = {2023-12-09},
	year = {2012}
}

@article{solove_myth_2021,
	title = {The Myth of the Privacy Paradox},
	volume = {89},
	url = {https://heinonline.org/HOL/P?h=hein.journals/gwlr89&i=15},
	pages = {1--51},
	number = {1},
	journal = {George Washington Law Review},
	author = {Solove, Daniel J.},
	urldate = {2023-12-09},
	year = {2021}
}

@misc{cate_failure_2006,
	location = {Rochester, {NY}},
	title = {The Failure of Fair Information Practice Principles},
	url = {https://papers.ssrn.com/abstract=1156972},
	number = {1156972},
    year={2006},
	author = {Cate, Fred H.},
	urldate = {2023-12-09},
	date = {2006},
	langid = {english},
}

@misc{noauthor_eur-lex_1992,
	title = {{EUR}-Lex - 32016R0679 - {EN} - {EUR}-Lex},
	url = {https://eur-lex.europa.eu/eli/reg/2016/679/oj},
	type = {European Parliament},
	urldate = {2023-12-12},
	date = {1992-10},
	langid = {english},
	note = {Doc {ID}: 32016R0679
Doc Sector: 3
Doc Title: Regulation ({EU}) 2016/679 of the European Parliament and of the Council of 27 April 2016 on the protection of natural persons with regard to the processing of personal data and on the free movement of such data, and repealing Directive 95/46/{EC} (General Data Protection Regulation) (Text with {EEA} relevance)
Doc Type: R
Usr\_lan: en},
    year={1992}	
}

@misc{noauthor_apec_2004,
	title = {{APEC} Privacy Framework},
	url = {https://www.apec.org/publications/2005/12/apec-privacy-framework},
	titleaddon = {Asia-Pacific Economic Cooperation},
	langid = {english},
    year={2004}
}

@misc{noauthor_privacy_1998,
	title = {Privacy Online: A Report to Congress 7},
	url = {https://www.ftc.gov/reports/privacy-online-report-congress},
	shorttitle = {Privacy Online},
	titleaddon = {Federal Trade Commission},
	urldate = {2023-12-12},
	date = {1998-06},
    year = {1998},
	langid = {english}
}

@misc{noauthor_oecd_1980,
	title = {{OECD} Privacy Principles},
	url = {http://oecdprivacy.org/},
	titleaddon = {Organisation for Economic Cooperation and Development},
	urldate = {2023-12-12},
	date = {1980-10},
    year = {1980}
}

@article{bohacek_protecting_2022_arxiv,
  title={Protecting President Zelenskyy against deep fakes},
  author={Bohacek, Matyas and Farid, Hany},
  journal={arXiv preprint arXiv:2206.12043},
  year={2022}
}

@article{bohacek_protecting_2022_pnas,
  title={Protecting world leaders against deep fakes using facial, gestural, and vocal mannerisms},
  author={Bohacek, Matyas and Farid, Hany},
  journal={Proceedings of the National Academy of Sciences},
  volume={119},
  number={48},
  pages={e2216035119},
  year={2022},
  publisher={National Acad Sciences}
}

@article{rajbahadur2021can,
  title={Can I use this publicly available dataset to build commercial AI software?--A Case Study on Publicly Available Image Datasets},
  author={Rajbahadur, Gopi Krishnan and Tuck, Erika and Zi, Li and Lin, Dayi and Chen, Boyuan and Ming, Zhen and German, Daniel M and others},
  journal={arXiv preprint arXiv:2111.02374},
  year={2021}
}

@article{elazar2023s,
  title={What's In My Big Data?},
  author={Elazar, Yanai and Bhagia, Akshita and Magnusson, Ian and Ravichander, Abhilasha and Schwenk, Dustin and Suhr, Alane and Walsh, Pete and Groeneveld, Dirk and Soldaini, Luca and Singh, Sameer and others},
  journal={arXiv preprint arXiv:2310.20707},
  year={2023}
}

@article{birhane2021multimodal,
  title={Multimodal datasets: misogyny, pornography, and malignant stereotypes},
  author={Birhane, Abeba and Prabhu, Vinay Uday and Kahembwe, Emmanuel},
  journal={arXiv preprint arXiv:2110.01963},
  year={2021}
}

@misc{wiki:datasets,
   author = "Wikipedia",
   title = "{List of datasets for machine-learning research} --- {W}ikipedia{,} The Free Encyclopedia",
   year = "2024",
   howpublished = {\url{http://en.wikipedia.org/w/index.php?title=List\%20of\%20datasets\%20for\%20machine-learning\%20research&oldid=1221075088}},
   note = "[Online; accessed 09-May-2024]"
 }

@article{lhoest2021datasets,
  title={Datasets: A community library for natural language processing},
  author={Lhoest, Quentin and del Moral, Albert Villanova and Jernite, Yacine and Thakur, Abhishek and von Platen, Patrick and Patil, Suraj and Chaumond, Julien and Drame, Mariama and Plu, Julien and Tunstall, Lewis and others},
  journal={arXiv preprint arXiv:2109.02846},
  year={2021}
}

@inproceedings{cosentino2016findings,
  title={Findings from GitHub: methods, datasets and limitations},
  author={Cosentino, Valerio and Luis, Javier and Cabot, Jordi},
  booktitle={Proceedings of the 13th International Conference on Mining Software Repositories},
  pages={137--141},
  year={2016}
}

@inproceedings{martinez2023towards,
  title={Towards understanding the interplay of generative artificial intelligence and the Internet},
  author={Mart{\'\i}nez, Gonzalo and Watson, Lauren and Reviriego, Pedro and Hern{\'a}ndez, Jos{\'e} Alberto and Juarez, Marc and Sarkar, Rik},
  booktitle={International Workshop on Epistemic Uncertainty in Artificial Intelligence},
  pages={59--73},
  year={2023},
  organization={Springer}
}

@article{denton2021genealogy,
  title={On the genealogy of machine learning datasets: A critical history of ImageNet},
  author={Denton, Emily and Hanna, Alex and Amironesei, Razvan and Smart, Andrew and Nicole, Hilary},
  journal={Big Data \& Society},
  volume={8},
  number={2},
  pages={20539517211035955},
  year={2021},
  publisher={SAGE Publications Sage UK: London, England}
}

@article{shope2021lawyer,
  title={Lawyer and judicial competency in the era of artificial intelligence: Ethical requirements for documenting datasets and machine learning models},
  author={Shope, Mark L},
  journal={Geo. J. Legal Ethics},
  volume={34},
  pages={191},
  year={2021},
  publisher={HeinOnline}
}

@inproceedings{deng2009imagenet,
  title={{ImageNet}: A large-scale hierarchical image database},
  author={Deng, Jia and Dong, Wei and Socher, Richard and Li, Li-Jia and Li, Kai and Fei-Fei, Li},
  booktitle={2009 {IEEE} conference on computer vision and pattern recognition},
  pages={248--255},
  year={2009},
  organization={IEEE}
}

@article{andreotta2022ai,
  title={AI, big data, and the future of consent},
  author={Andreotta, Adam J and Kirkham, Nin and Rizzi, Marco},
  journal={Ai \& Society},
  volume={37},
  number={4},
  pages={1715--1728},
  year={2022},
  publisher={Springer}
}

@article{carter2020regulation,
  title={Regulation and ethics in artificial intelligence and machine learning technologies: Where are we now? Who is responsible? Can the information professional play a role?},
  author={Carter, Denise},
  journal={Business Information Review},
  volume={37},
  number={2},
  pages={60--68},
  year={2020},
  publisher={SAGE Publications Sage UK: London, England}
}

@inproceedings{chu2024protect,
  title={How to Protect Copyright Data in Optimization of Large Language Models?},
  author={Chu, Timothy and Song, Zhao and Yang, Chiwun},
  booktitle={Proceedings of the AAAI Conference on Artificial Intelligence},
  volume={38},
  number={16},
  pages={17871--17879},
  year={2024}
}

@article{levendowski2018copyright,
  title={How copyright law can fix artificial intelligence's implicit bias problem},
  author={Levendowski, Amanda},
  journal={Wash. L. Rev.},
  volume={93},
  pages={579},
  year={2018},
  publisher={HeinOnline}
}

@article{arora2023value,
  title={The value of standards for health datasets in artificial intelligence-based applications},
  author={Arora, Anmol and Alderman, Joseph E and Palmer, Joanne and Ganapathi, Shaswath and Laws, Elinor and McCradden, Melissa D and Oakden-Rayner, Lauren and Pfohl, Stephen R and Ghassemi, Marzyeh and McKay, Francis and others},
  journal={Nature Medicine},
  volume={29},
  number={11},
  pages={2929--2938},
  year={2023},
  publisher={Nature Publishing Group US New York}
}

@inproceedings{kusak2022quality,
  title={Quality of data sets that feed AI and big data applications for law enforcement},
  author={Kusak, Martyna},
  booktitle={ERA Forum},
  volume={23},
  number={2},
  pages={209--219},
  year={2022},
  organization={Springer}
}

@article{khan2022subjects,
  title={The subjects and stages of ai dataset development: A framework for dataset accountability},
  author={Khan, Mehtab and Hanna, Alex},
  journal={Ohio St. Tech. LJ},
  volume={19},
  pages={171},
  year={2022},
  publisher={HeinOnline}
}

@article{heger2022understanding,
  title={Understanding machine learning practitioners' data documentation perceptions, needs, challenges, and desiderata},
  author={Heger, Amy K and Marquis, Liz B and Vorvoreanu, Mihaela and Wallach, Hanna and Wortman Vaughan, Jennifer},
  journal={Proceedings of the ACM on Human-Computer Interaction},
  volume={6},
  number={CSCW2},
  pages={1--29},
  year={2022},
  publisher={ACM New York, NY, USA}
}

@article{longpre2023data,
  title={The data provenance initiative: A large scale audit of dataset licensing \& attribution in ai},
  author={Longpre, Shayne and Mahari, Robert and Chen, Anthony and Obeng-Marnu, Naana and Sileo, Damien and Brannon, William and Muennighoff, Niklas and Khazam, Nathan and Kabbara, Jad and Perisetla, Kartik and others},
  journal={arXiv preprint arXiv:2310.16787},
  year={2023}
}

@article{verhulst2023urgent,
  doi = {10.48558/TDS9-6Y22},
  url = {https://ssir.org/articles/entry/the_urgent_need_to_reimagine_data_consent},
  author = {Verhulst,  Stefaan G. and Sandor,  Laura and Stamm,  Julia},
  language = {en},
  title = {The Urgent Need to Reimagine Data Consent},
  publisher = {Stanford Social Innovation Review},
  year = {2023}
}

@article{labastida2020licensing,
  title={Licensing FAIR data for reuse},
  author={Labastida, Ignasi and Margoni, Thomas},
  journal={Data Intelligence},
  volume={2},
  number={1-2},
  pages={199--207},
  year={2020},
  publisher={MIT Press One Rogers Street, Cambridge, MA 02142-1209, USA journals-info~…}
}

@misc{Li_2023, title={Data scraping makes AI systems possible, but at whose expense?}, url={https://www.techpolicy.press/data-scraping-makes-ai-systems-possible-but-at-whose-expense/}, journal={Tech Policy Press}, publisher={Tech Policy Press}, author={Li, Hanlin}, year={2023}, month={Oct}}

@article{zirpoli2023generative,
  title={Generative artificial intelligence and copyright law},
  author={Zirpoli, Christopher T},
  year={2023}
}

@article{zhang2023text,
  title={Text-to-image diffusion model in generative ai: A survey},
  author={Zhang, Chenshuang and Zhang, Chaoning and Zhang, Mengchun and Kweon, In So},
  journal={arXiv preprint arXiv:2303.07909},
  year={2023}
}

@inproceedings{kondo2023mva2023,
  title={Mva2023 small object detection challenge for spotting birds: Dataset, methods, and results},
  author={Kondo, Yuki and Ukita, Norimichi and Yamaguchi, Takayuki and Hou, Hao-Yu and Shen, Mu-Yi and Hsu, Chia-Chi and Huang, En-Ming and Huang, Yu-Chen and Xia, Yu-Cheng and Wang, Chien-Yao and others},
  booktitle={2023 18th International Conference on Machine Vision and Applications (MVA)},
  pages={1--11},
  year={2023},
  organization={IEEE}
}

@inproceedings{lin2014microsoft,
  title={Microsoft coco: Common objects in context},
  author={Lin, Tsung-Yi and Maire, Michael and Belongie, Serge and Hays, James and Perona, Pietro and Ramanan, Deva and Doll{\'a}r, Piotr and Zitnick, C Lawrence},
  booktitle={Computer Vision--ECCV 2014: 13th European Conference, Zurich, Switzerland, September 6-12, 2014, Proceedings, Part V 13},
  pages={740--755},
  year={2014},
  organization={Springer}
}

@inproceedings{jafarian2021learning,
  title={Learning high fidelity depths of dressed humans by watching social media dance videos},
  author={Jafarian, Yasamin and Park, Hyun Soo},
  booktitle={Proceedings of the IEEE/CVF Conference on Computer Vision and Pattern Recognition},
  pages={12753--12762},
  year={2021}
}

@inproceedings{rombach2022high,
  title={High-resolution image synthesis with latent diffusion models},
  author={Rombach, Robin and Blattmann, Andreas and Lorenz, Dominik and Esser, Patrick and Ommer, Bj{\"o}rn},
  booktitle={Proceedings of the IEEE/CVF conference on computer vision and pattern recognition},
  pages={10684--10695},
  year={2022}
}

\appendix

\label{appendix:crs-scale}

\begin{figure*}[t]

\section{A. CRS Scale Visuals}
\centering

  \subcaptionbox*{}[.45\linewidth]{%
   \includegraphics[width=0.46\textwidth]{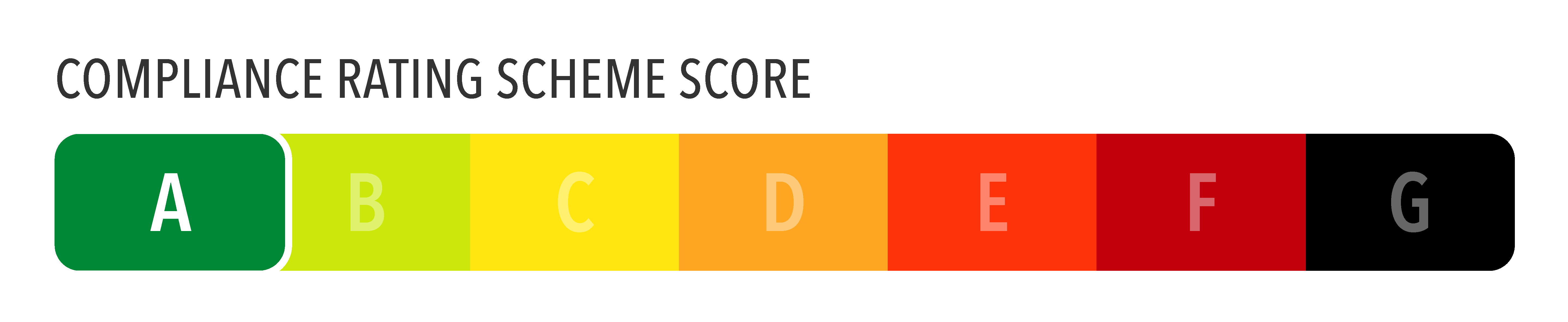}
  }%
  \hfill
  \subcaptionbox*{}[.45\linewidth]{%
    \includegraphics[width=0.46\textwidth]{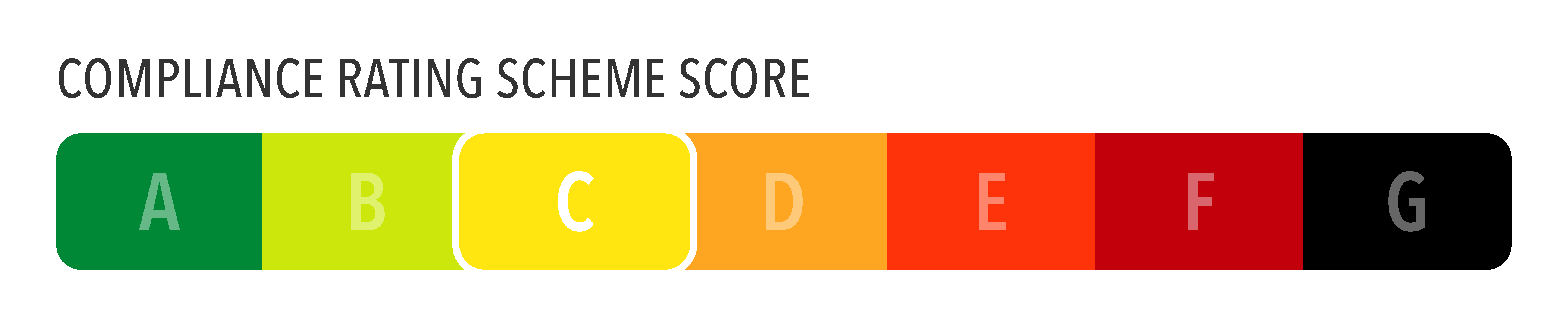}
  }

  \vspace{-0.5cm}
  
  \caption{Proposed design interface for "A" and "C" score on the CRS scale}
  \label{fig:CRS-A}
\end{figure*}

\label{appendix:additionalmockups}

\begin{figure*}

\section{B. Schematic Overviews of the \textit{DatasetSentinel} Use Cases}

  \centering
  \includegraphics[width=0.95\textwidth]{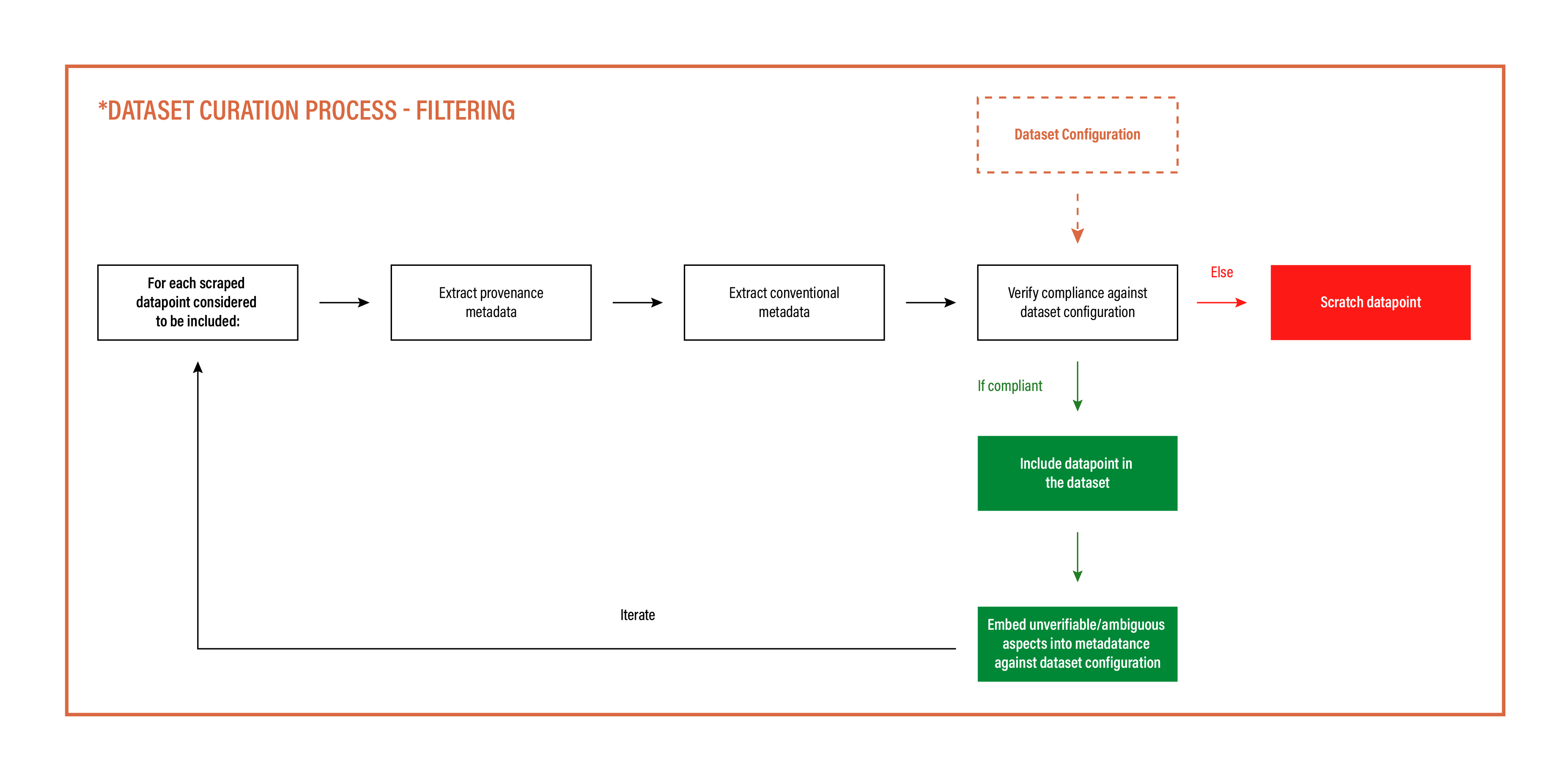}
  
  \caption{A schematic overview of DatasetSentinel's use case within the dataset curation stage of the dataset lifecycle.}
  \label{fig:use-case-1}
\end{figure*}

\begin{figure*}
  \centering
  \includegraphics[width=0.95\textwidth]{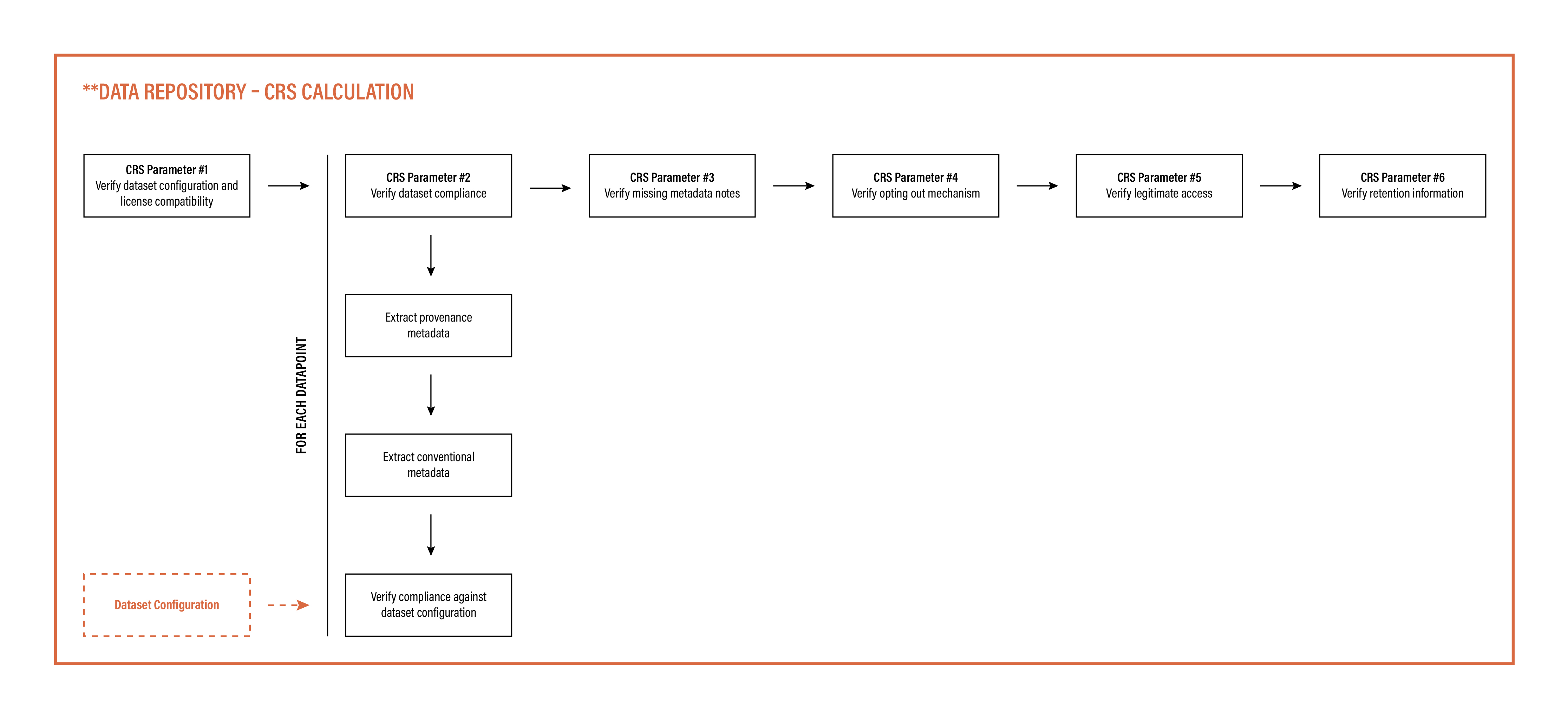}
  
  \caption{A schematic overview of CRS' use case within the dataset repository stage of the dataset lifecycle.}
  \label{fig:use-case-2}
\end{figure*}

\begin{figure*}[h]

\section{C. Additional CRS Mockups}
\label{appendix:additionalmockups}

\vspace{1cm}

  \centering
  \includegraphics[width=0.8\textwidth]{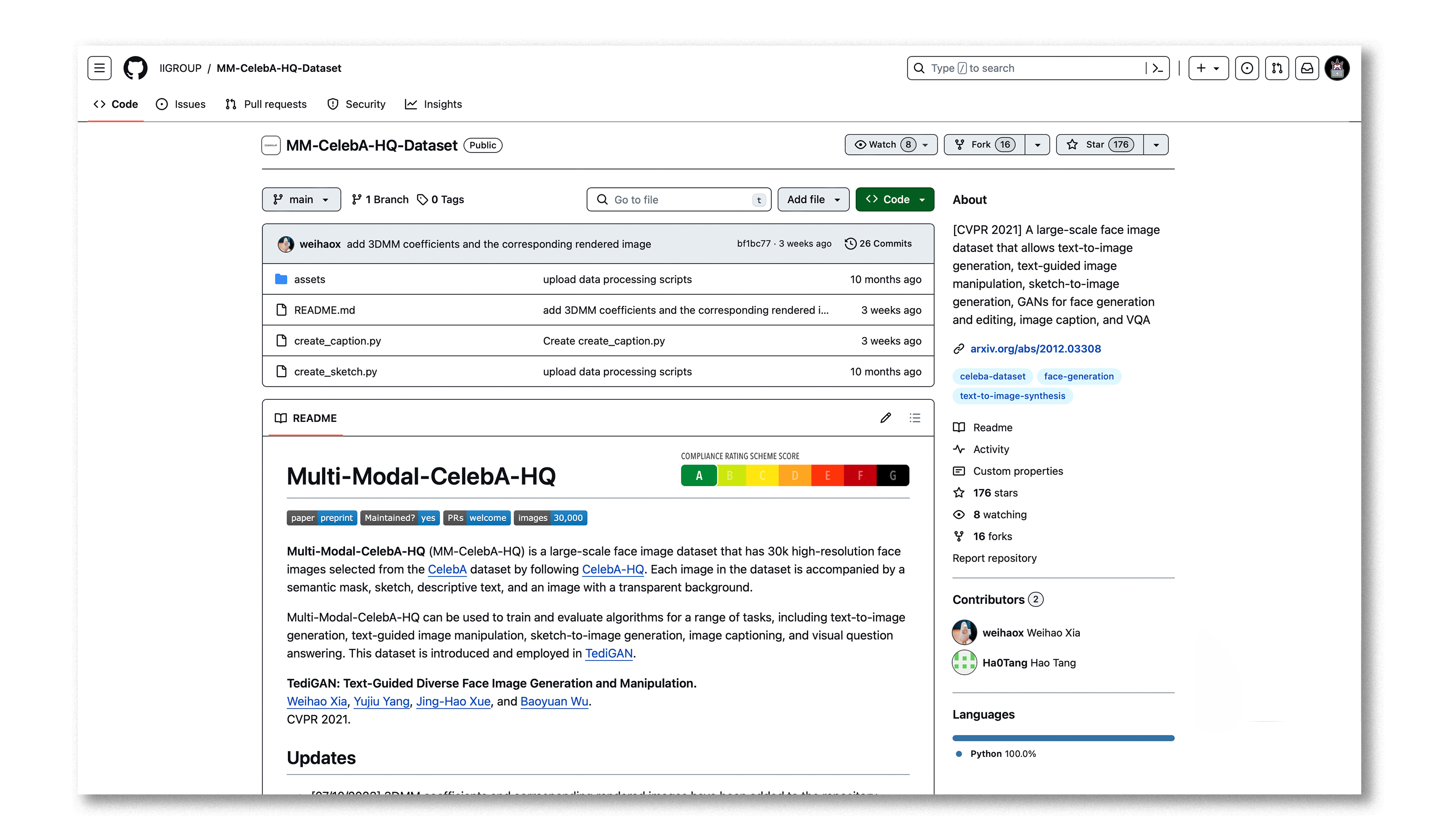}
  
  \caption{A fictitious "A" CRS score mock-up of a random GitHub dataset}
  \label{fig:mockup-github}
\end{figure*}

\begin{figure*}[]
  \centering
  \includegraphics[width=0.8\textwidth]{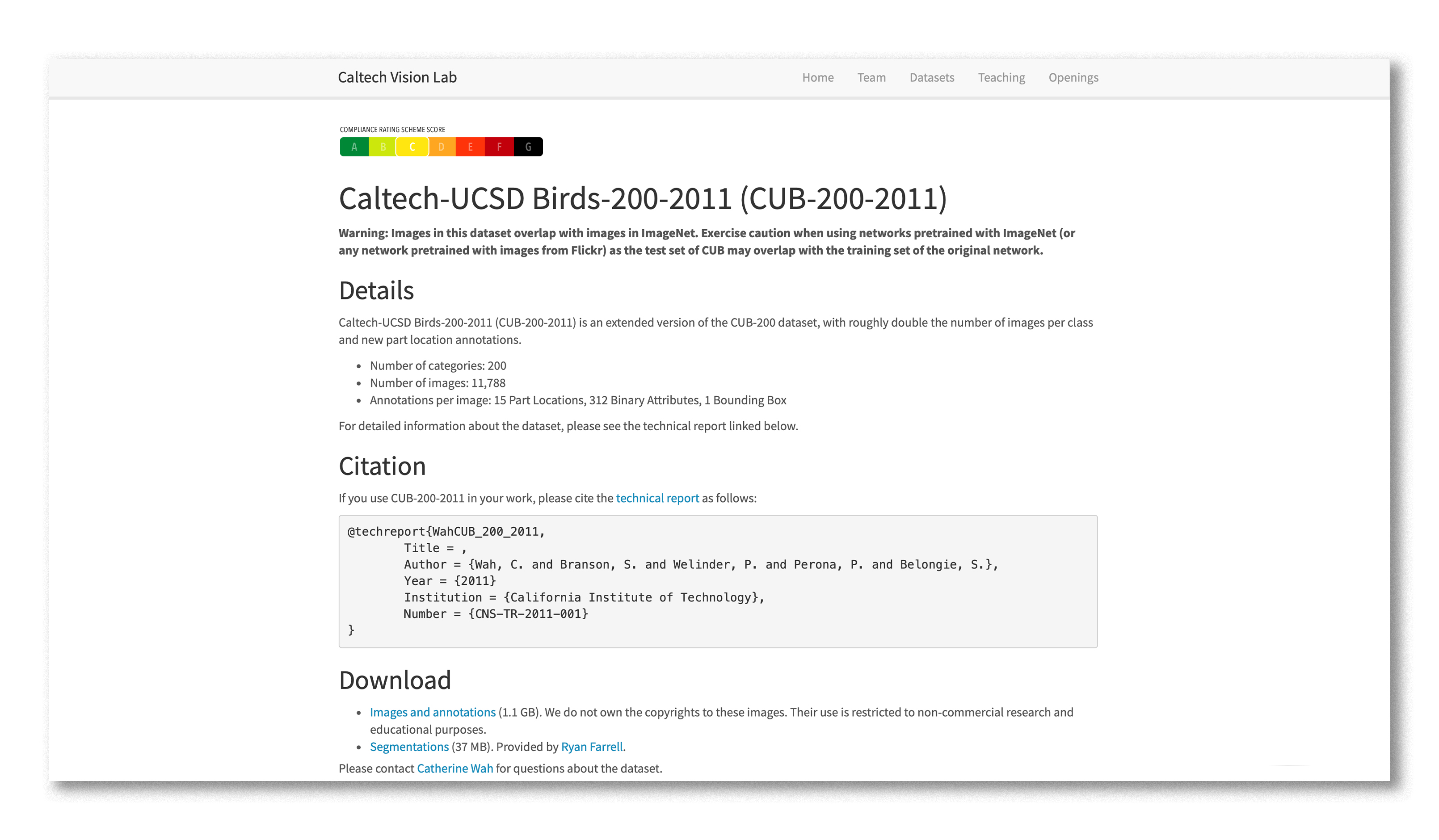}
  
  \caption{A fictitious CRS "C" score mock-up of a random academic repository dataset}
  \label{fig:mockup-lab-website}
\end{figure*}

\begin{figure*}
  \centering
  \includegraphics[width=0.8\textwidth]{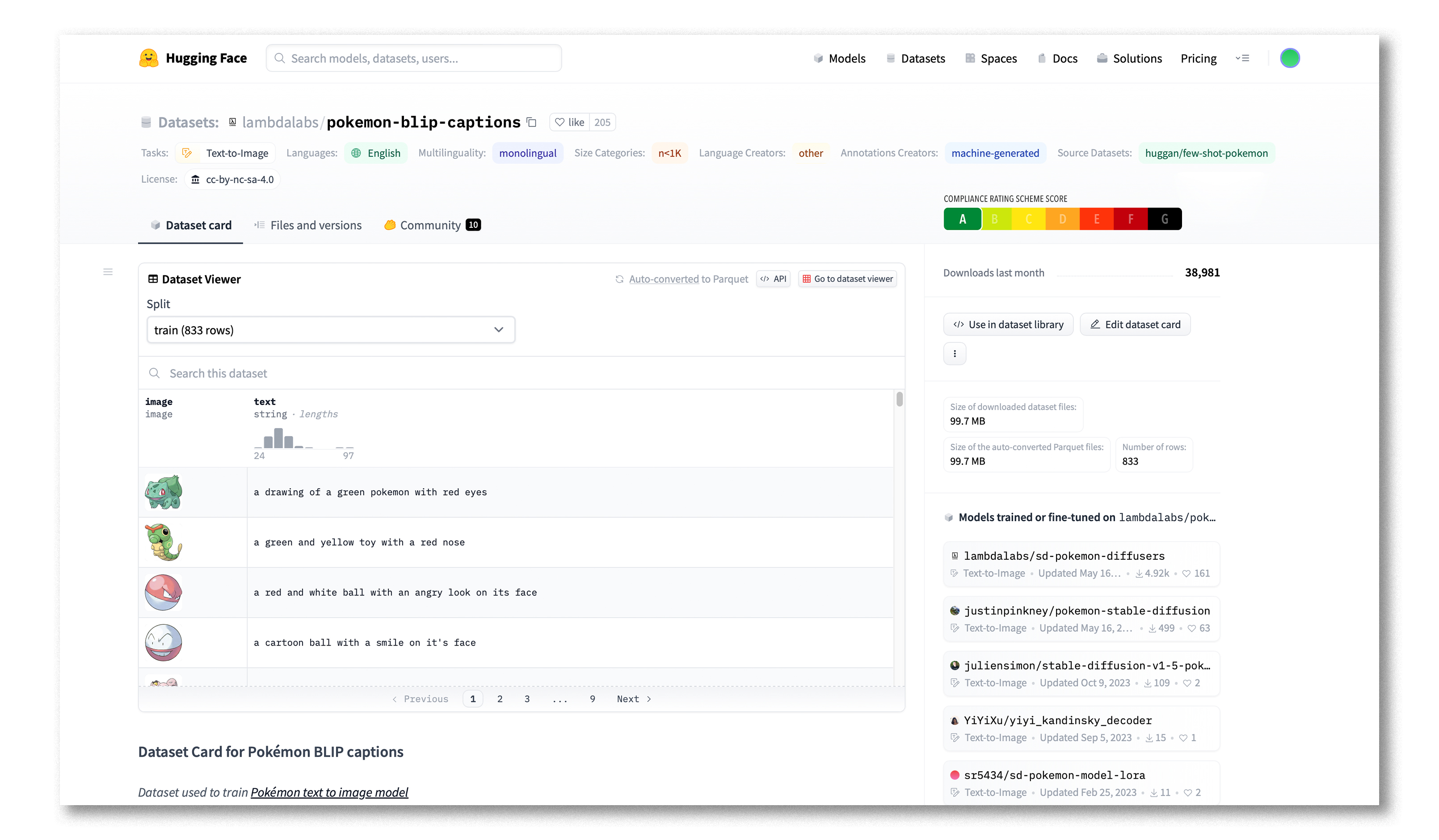}
  
  \caption{A fictitious CRS "A" score mock-up of a random Hugging Face dataset}
  \label{fig:mockup-hf}
\end{figure*}

\end{document}